\documentclass[sigconf]{acmart}
\AtBeginDocument{%
  }

\usepackage{graphicx} 
\usepackage{booktabs} 
\usepackage{multirow} 
\usepackage{amsmath}  
\usepackage{subcaption} 
\usepackage{algorithm}    
\usepackage{algorithmic}
\usepackage{color}
\usepackage{xcolor}
\usepackage{amsmath}

\usepackage[encapsulated]{CJK}
\usepackage{makecell}
\newcommand{\chinese}[1]{\begin{CJK}{UTF8}{gbsn} #1 \end{CJK}}
\definecolor{cyan}{RGB}{154,201,219}
\definecolor{red}{RGB}{255,0,0}
\definecolor{flesh}{RGB}{192,0,0}
\definecolor{orange}{RGB}{237,125,49}
\usepackage{soul}
\usepackage{comment}
\usepackage{stackengine}

\usepackage{pifont}
\usepackage{wrapfig}
\usepackage{threeparttable}
\usepackage{tcolorbox}
\usepackage{verbatim}
\usepackage{listings}
\usepackage{courier}
\tcbuselibrary{listings, breakable}

\lstdefinelanguage{plainjson}{
  basicstyle=\ttfamily\fontsize{5}{6},
  breaklines=true,
  showstringspaces=false,
  escapeinside={(*@}{@*)} 
}

\renewcommand\footnotetextcopyrightpermission[1]{} 

\setcopyright{acmlicensed}
\copyrightyear{2025}
\acmYear{2025}
\acmDOI{XXXXXXX.XXXXXXX}
\acmConference[KDD '26]{In Proceedings of the 32th ACM SIGKDD Conference on Knowledge Discovery and Data Mining }{August 9-13, 2026}{Jeju, Korea}

\acmISBN{978-1-4503-XXXX-X/2025/08}

\begin{document}

\title{Automated Annotation of Privacy Information in User
Interactions with Large Language Models}

\author{Hang Zeng}
\affiliation{%
  \institution{Shanghai Jiao Tong University}
  \city{Shanghai}
  \country{China}}
\email{nidhogg@sjtu.edu.cn}

\author{Xiangyu Liu}
\affiliation{%
  \institution{WeChat Tencent}
  \city{Beijing}
  \country{China}}
\email{xiangyuliu@tencent.com}

\author{Yong Hu}
\affiliation{%
  \institution{WeChat Tencent}
  \city{Beijing}
  \country{China}}
\email{rightyonghu@tencent.com}

\author{Chaoyue Niu}
\authornote{Chaoyue Niu is the corresponding author.}
\affiliation{%
  \institution{Shanghai Jiao Tong University}
  \city{Shanghai}
  \country{China}
}
\email{rvince@sjtu.edu.cn}

\author{Fan Wu}
\affiliation{%
  \institution{Shanghai Jiao Tong University}
  \city{Shanghai}
  \country{China}
}
\email{wu-fan@sjtu.edu.cn}

\author{Shaojie Tang}
\affiliation{%
  \institution{University at Buffalo}
  \city{New York}
  \country{United States}
}
\email{shaojiet@buffalo.edu}

\author{Guihai Chen}
\affiliation{%
  \institution{Shanghai Jiao Tong University}
  \city{Shanghai}
  \country{China}
}
\email{gchen@cs.sjtu.edu.cn}

\renewcommand{\shortauthors}{Zeng et al.}

\begin{abstract}
Users interacting with large language models (LLMs) under their real identifiers often unknowingly risk disclosing private information. Automatically notifying users whether their queries leak privacy and which phrases leak what private information has therefore become a practical need. Existing privacy detection methods, however, were designed for different objectives and application domains, typically tagging personally identifiable information (PII) in anonymous content, which is insufficient in real-name interaction scenarios with LLMs. In this work, to support the development and evaluation of privacy detection models for LLM interactions that are deployable on local user devices, we construct a large-scale multilingual dataset with 249K user queries and 154K annotated privacy phrases. In particular, we build an automated privacy annotation pipeline with strong LLMs to automatically extract privacy phrases from dialogue datasets and annotate leaked information. We also design evaluation metrics at the levels of privacy leakage, extracted privacy phrase, and privacy information. We further establish baseline methods using light-weight LLMs with both tuning-free and tuning-based methods, and report a comprehensive evaluation of their performance. Evaluation results reveal a gap between current performance and the requirements of real-world LLM applications, motivating future research into more effective local privacy detection methods grounded in our dataset\footnote{The code and dataset are public at \href{https://github.com/Nidryen-zh/PrivacyAnnotation}{https://github.com/Nidryen-zh/PrivacyAnnotation} and \href{https://huggingface.co/datasets/Nidhogg-zh/Interaction_Dialogue_with_Privacy}{https://huggingface.co/datasets/Nidhogg-zh/Interaction$\_$Dialogue$\_$with$\_$Privacy}.}.
\end{abstract}

\ccsdesc[500]{Security and privacy}
\ccsdesc[500]{Security and privacy~Human and societal aspects of security and privacy}
\ccsdesc[300]{Security and privacy~Privacy protections}
\ccsdesc[500]{Computing methodologies}
\ccsdesc[500]{Computing methodologies~Natural language processing}
\ccsdesc[300]{Computing methodologies~Discourse, dialogue and pragmatics}

\keywords{LLM Interaction, Privacy Detection Dataset}



\maketitle

\section{Introduction}
\begin{figure}[!t]
    \centering
    \includegraphics[width=0.85\linewidth]{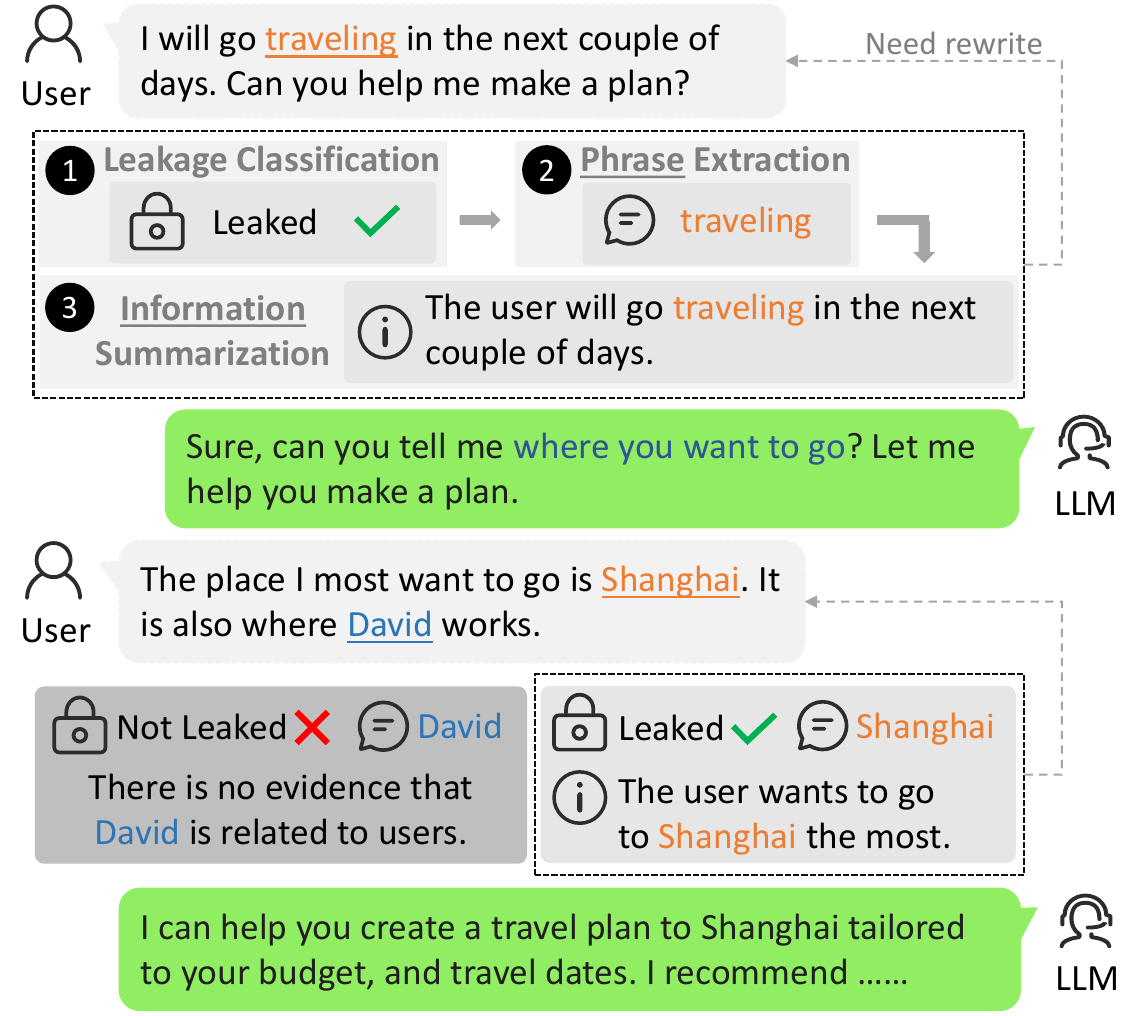}
    \caption{An illustration of privacy exposure and annotation during a user's interactions with LLM.}\label{fig:case}
\end{figure}

Large language models (LLMs) have demonstrated immense potential in revolutionizing user interactions with artificial intelligence. However, during cloud-based LLM interactions, users often unintentionally expose privacy to service providers \cite{privacyconcern, survey, mope, assessingprivacy, LPLM, beyond, decoding, trust_no_bot}. A case study of the ShareGPT corpus \cite{sharegpt} reveals that nearly one-third of dialogues contain privacy leaks. One such instance of exposing the user's intents and preferences is illustrated in Figure \ref{fig:case}. Even more concerning, LLM service providers typically require access to users’ or organizations’ identity information, exposing unique identifiers before any interaction occurs. This real-identity interaction setting significantly increases the risk of privacy information disclosure. Therefore, proactively alerting users to potential privacy exposure during interactions with LLMs has become an urgent and practical need. When a user launches a query, whether the query contains private information or not, should be informed. If so, the specific phrases in the query responsible for the privacy leakage need to be annotated. To further assist users who may still be uncertain, a concise explanation of what private information will be exposed by each annotated phrase should be provided. These notifications can help users quickly understand the nature of privacy leakage and take appropriate action, such as rewriting the query.

Previous research for privacy protection across different domains, including medical privacy \cite{medical1, medical2, medical3}, code privacy \cite{finding}, and employment information \cite{Multilingual, occupational}, primarily focuses on detecting privacy risks by identifying private entities that reveals personal identifiable information (PII) in anonymous online content or documents \cite{selfdisclosures, entity2, synthetic}. However, these scenarios differ significantly from privacy detection in LLM interactions. By authenticating through user login, queries posed to LLMs are linked to a specific user's real identifier. It is important to detect privacy information related to the current user from queries.
On the other hand, if users disclose private information about unrelated third parties, such information will not be extracted since it does not constitute a harmful breach of privacy for the current user. Additionally, many existing studies are constrained to a limited set of topics or categories \cite{ner1, anonymisation, categorylimit1, categorylimit2}. In contrast, privacy information in LLM interactions is much more diverse and cannot easily be classified into a limited set of types.

To enable privacy detection for real-time user queries during the interactions, it is desired to deploy an efficient model on the user side for automatic local detection before sending queries to cloud-based LLMs. The development and evaluation of such a model require large-scale datasets to ensure good generalization ability. However, no such comprehensive datasets currently exist for LLM interactions. Existing datasets were manually labeled and limited to particular domains or categories, with only a few thousand entries \cite{tab, selfdisclosures, i2b2, cegs, codealltag}. To address this, we design an automated, scalable privacy annotation pipeline to extract privacy phrases and summarize the corresponding privacy information for user dialogues. With the assistance of GPT-4o, our pipeline operates through four systematic steps: (1) privacy leakage classification, which filters the input to determine whether the user's query leaks privacy; (2) extensive privacy categories extraction, which identifies privacy categories to guide the extraction of phrases; (3) privacy phrase extraction, where privacy phrases are extracted from queries and filtered; and (4) privacy information annotation, which summarizes details for each privacy phrase. Based on the automated pipeline, we extract privacy phrases and annotate privacy information for dialogue datasets in different languages, containing 97,659 user queries, 85,320 phrases with corresponding information for the English portion, and 151,988 user queries, 68,910 phrases with information for the Chinese portion. We also conduct human evaluation, demonstrating that the results obtained through our pipeline are highly consistent with those manually annotated. Compared to directly prompting LLM to extract privacy phrases \cite{ghost}, our pipeline produces more accurate and diverse annotations. 

The constructed dataset can serve not only as a benchmark for evaluation but also as a valuable source of training data for local privacy detection methods. 
For evaluation, we design a hierarchical set of metrics at query, phrase, and information levels, each corresponding to tasks of increasing granularity in privacy detection. Specifically, we use accuracy for privacy leakage classification, which determines whether a query reveals private information. For privacy phrase extraction, which identifies sensitive phrases within queries, and privacy information summarization, which captures the specific private information leaked, we design precision, recall, and F1 scores to assess performance. These metrics quantify the differences between the extracted phrases or generated privacy information and the ground truth, effectively evaluating both missed and incorrect extractions or summaries.
We also set up tuning-free baselines, including prompting and in-context learning with locally deployable, light-weight LLMs, as well as supervised fine-tuning baselines. Benchmark results show that the fine-tuned 1B model with our dataset outperforms the directly prompted 72B model.

We summarize the key contributions as follows: 
(1) We identify a new application requirement of privacy detection for real-name user interactions with LLMs;
(2) we build the automated privacy annotation pipeline to classify privacy leakage, extract privacy phrases and information, resulting in the first large-scale dataset for privacy detection in real-name interactions. We also design new metrics for benchmark testing; and  
(3) we report the performance of local privacy detection baselines, which, despite their progress, still fall short of meeting practical requirements. This highlights the need for further research into more effective methods grounded in our dataset. Specifically, the best-performing fine-tuned model achieves 87.6\% accuracy in privacy leakage classification, a 74.3\% F1 score in privacy phrase extraction, and a 44.7\% F1 score in privacy information summarization.

\section{Related Work}
\textbf{Privacy Risks of LLMs.}
Recent work constructs the corpora with online personal information or user interaction history \cite{privacysurvey2, finetuneingleak} to train LLMs \cite{characterllm, exploitingpersona}, and enhance inference ability \cite{learningretrieval, hallucination}, resulting in the issue of memorization \cite{memorization, privacysurvey1, ProPILE, extractingtraining}, which recalls the personal sequences from LLM corpora. 
Existing works solve this problem on the LLM service provider side, including data pre-processing approaches (e.g., sanitizing training data \cite{sanitizing} and reducing the frequency of private data appearing \cite{Deduplicating1, Deduplicating2}), and privacy-preserving training methods (e.g., differentially private training \cite{differential1, differential2, differential3}). In contrast, we take a different angle for privacy risks of LLMs by locally detecting privacy in the user interactions with LLMs to prevent private information from being uploaded to LLM service providers, solving the problem at the root of the user side.

\textbf{Privacy Protection.} 
Most previous work focuses on privacy detection for specific documents or context shared by users on the Internet \cite{semantics, privacypreserving}, thereby removing PII to achieve anonymization \cite{entity1, entity2, synthetic}. 
However, privacy detection in user interactions with LLMs, where users have already authenticated their identifiers, is fundamentally different from these scenarios. 
The target application scenario of this work is to detect any newly leaked private data related to the user after a query is made, extending beyond PII.
Additionally, existing approaches are typically constrained to specific topics or categories \cite{anonymisation, personalinformation}, such as medical privacy \cite{medical1, medical2, medical3, medical4}, personal employment information \cite{Multilingual, occupational}, online comments \cite{selfdisclosures}, and code privacy \cite{finding}. In contrast, the types of privacy information in LLM interactions are not limited. Furthermore, most existing privacy detection models are based on discriminative models, treating the detection process as a sequence tagging task \cite{selfdisclosures, ner1, entity3}. They can not generate specific privacy information to remind the user. 
MemoAnalyzer \cite{ghost} directly prompts LLMs to identify open privacy information without predefining categories, but the detection performance remains suboptimal. 
Besides privacy detection, PAPILLON \cite{papillon} introduces a method that prompts a local model to rewrite user queries before they are uploaded to prevent privacy leakage. However, its protection remains limited to PII.

\begin{table}[!t]
\centering
\caption{The comparison with the existing datasets.}\label{table:dataset_compare}
\resizebox{1\linewidth}{!}{
    \begin{tabular}{l|c|c|c|c|c}
    \toprule
        Dataset                        & Target  & Labeling  & Data Size  & Information            & Language\\
    \midrule
        I2B2/UTHealth \cite{i2b2}         & Personal Health   & Manual & 1,304     & $\times$        & EN \\
        CodE ALLTAG \cite{codealltag}     & Email             & Manual & 2,309     & $\times$        & DE \\
        TAB \cite{tab}                    & Court Judgments   & Manual & 1,268     & $\times$        & EN \\
        Reddit-Self-Disclosures \cite{selfdisclosures}  & Comments          & Manual & 2,415     & $\times$        & EN \\
        SynthPAI \cite{synthpai}          & Comments          & Manual & 7,823     & $\times$        & EN \\
    \midrule
        Ours                            & LLM Interactions  & Automated  & 249,683  & $\surd$        & EN, ZH \\
    \bottomrule
    \end{tabular}
    }
\end{table}

\textbf{Privacy Datasets.}
Existing datasets for privacy detection are limited to specific domains and there is no large-scale dataset for real-name LLM interactions. For instance, PrivacyBot \cite{privacybot} provides a sentence-level privacy classification dataset, while I2B2/UTHealth \cite{i2b2} and CEGS N-GRID \cite{cegs} focus on privacy entity recognition in the personal health domain. CodE ALLTAG \cite{codealltag} targets privacy entity recognition in German emails, TAB \cite{tab} annotates private entities in court judgments, and JobStack \cite{jobstack} specializes in identifying such entities in job postings. 
SynthPAI \cite{synthpai} annotates simulated comments with inferable personal attributes, such as age and gender. 
Additionally, these datasets face challenges of data volume, and they are manually labeled. 
In contrast, our dataset focuses on LLM interactions and is annotated automatically, resulting in a significantly larger scale compared to existing datasets and encompassing a broader range of privacy. Furthermore, it offers detailed information for each extracted privacy phrase. Table \ref{table:dataset_compare} presents a comparison between our dataset and existing datasets.

\section{Problem Formulation}\label{sec:setup}

\textbf{Privacy in LLM Interactions.} We focus on the scenario where the user interacts with LLMs by submitting real-name text queries. 
The LLM service providers are honest-but-curious, which implies that they may attempt to infer privacy information from queries and do not actively launch attacks.
Different from PII, which refers to data that can identify an individual in an anonymous text, we define privacy information in LLM interactions as the service provider's knowledge about user-related private data or personal details, before and after the user poses a query with a real, unique identifier. Specifically, the private data in LLM interactions refers to the data that, if exposed, could negatively impact the personal, social, or professional aspects of the current user or the user’s related people, such as family, friends, or close associates.
This definition of privacy information leakage is consistent with the principle of differential privacy, which considers the difference in the query output with and without each individual data record. Typical privacy information includes, but is not limited to, sensitive preferences, intentions, and opinions. 
We also define privacy phrases as portions of user queries that leak privacy information. The privacy phrase for the user's first query in Figure \ref{fig:case} is ``traveling'', and the corresponding privacy information is ``The user will go traveling in the next couple of days.''
We emphasize that the privacy information in LLM interactions should be related to the current user who poses the query with direct evidence. For example, as shown in Figure \ref{fig:case}, the statement ``Shanghai is also where David works.'' leaks privacy information about David, but it does not leak the privacy information about the current user, since there is no direct evidence indicating David is related to the user. 

\begin{figure}[!t]
    \centering
        \includegraphics[width=0.9\linewidth]{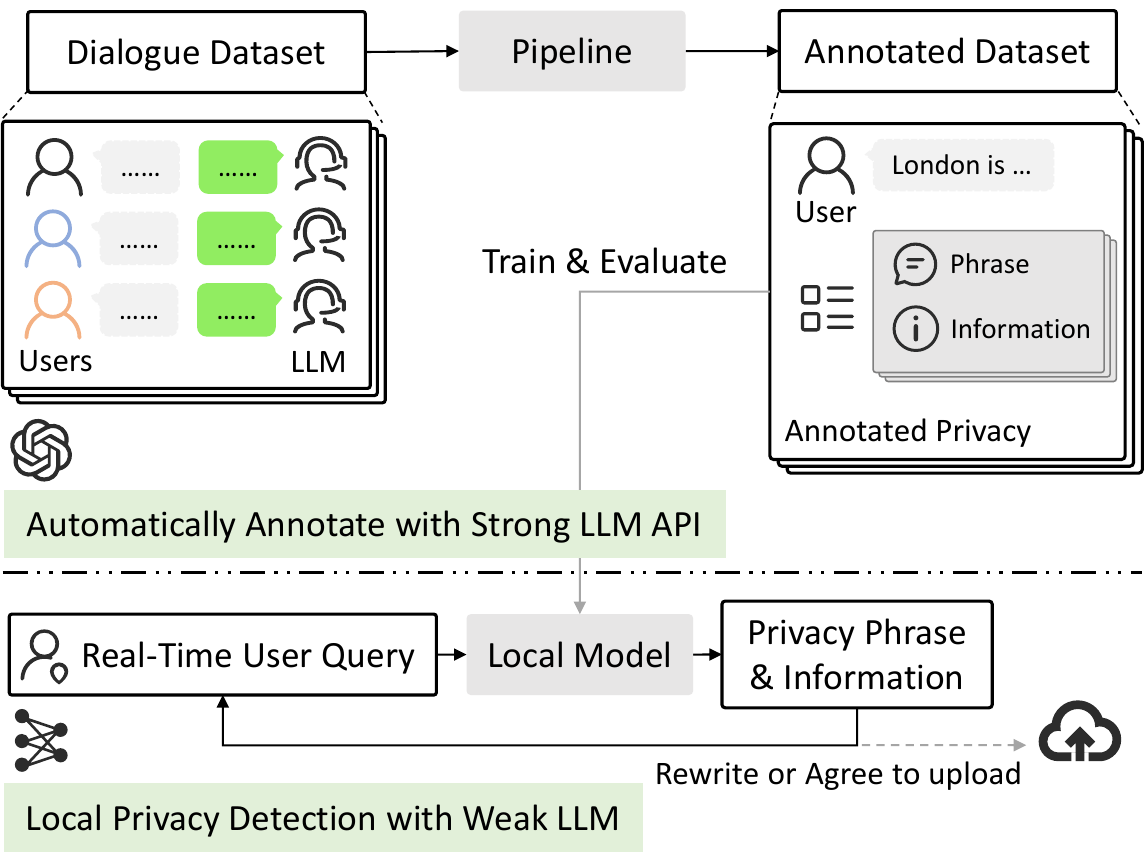}
        \caption{Automated privacy annotation pipeline with strong LLM and privacy detection module with light-weight LLM.}\label{fig:pipeline_and_local}
\end{figure}

\textbf{Automated Privacy Annotation Requirement.}
As shown in Figure \ref{fig:pipeline_and_local}, to implement privacy leakage alerting on the user side, it is necessary to first build an automated privacy annotation pipeline based on a cloud-side LLM API to annotate dialogue datasets with privacy phrases and specific privacy information. 
Such a dataset can serve as an evaluation benchmark for downstream applications and provide training data for optimization. Local privacy detection methods with light-weight LLM are then designed to detect privacy from user real-time queries before they are sent to service providers, and they should be deployed locally on the user's device or be a request proxy built by the user's organization.
Formally, we let $D$ denote the user dialogue with LLM and $q$ denote a query from $D$. The goal is to extract the privacy phrase $\mathcal{P}_q=\{p_1, ..., p_m\}$ and summarize privacy information $\mathcal{I}_q=\{I_1, ..., I_m\}$, where $m$ is the number of the extracted phrases, $p_j$ is a segment of $q$ that discloses private or sensitive information about the user, and $I_j$ is a statement summarizing the specific details revealed by $p_j$.

\begin{figure*}[!t]
  \centering
         \centering
         \includegraphics[width=1\textwidth]{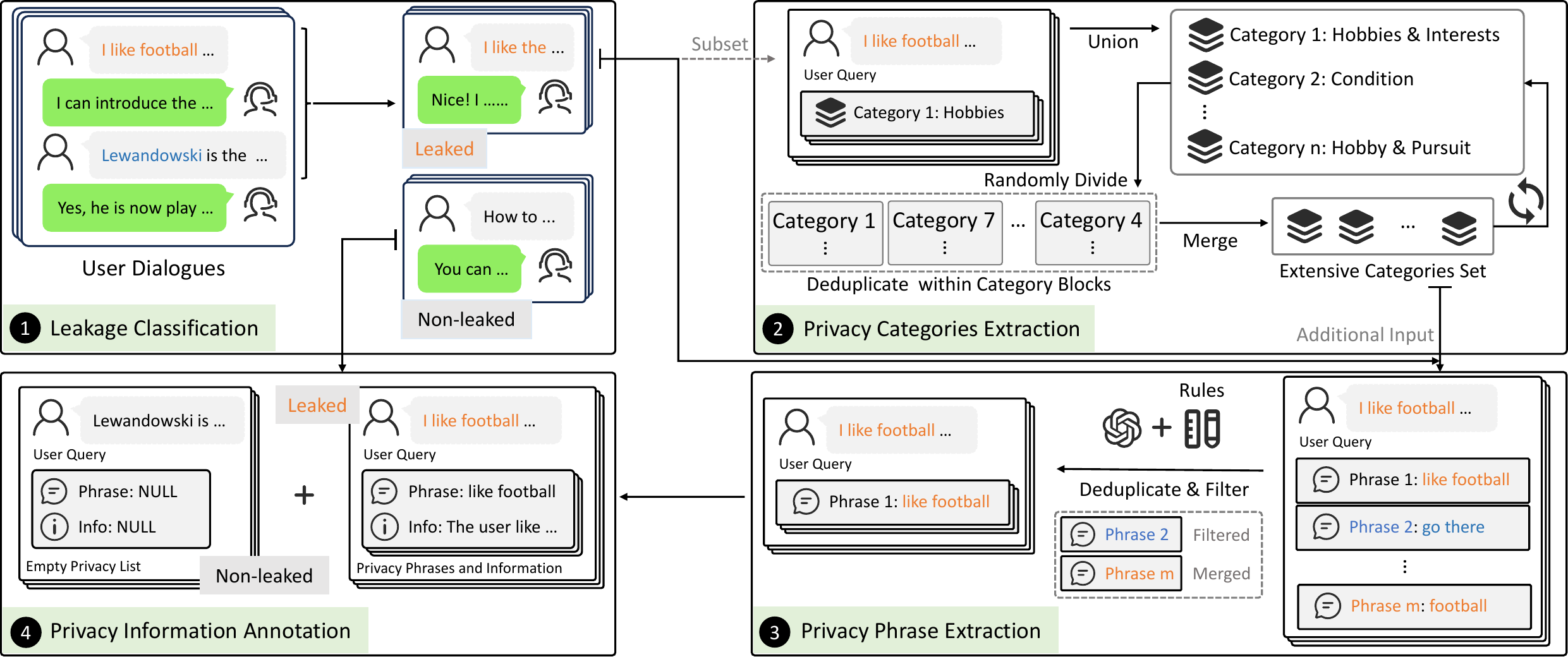}
         \label{fig:trad}
    \caption{Overview of our automated pipeline to extract privacy phrase and annotate privacy information over dialogue datasets: \ding{202} determining whether the input leaks privacy; \ding{203} extracting privacy categories from user queries that leaks privacy; \ding{204} extracting privacy phrases from user queries with the help of privacy categories; and \ding{205} annotating privacy information for each phrase.}\label{fig:framework}
    
\end{figure*}

\section{Automated Privacy Annotation Pipeline for Dialogue Datasets}

We design an automated four-step pipeline based on the strong LLM API to extract privacy phrases and annotate privacy information for dialogue datasets.

\subsection{Pipeline Rationale}
A straightforward approach to extract privacy information is directly prompting GPT-4o. However, this method may force GPT-4o to extract information from samples that do not contain any privacy leaks, which results in a decrease in precision. 
Furthermore, the original understanding of privacy by GPT-4o revolves around the idea of whether a privacy phrase can reveal a user's identifiable information. While this is intuitive for identifying PII, it does not align with our intended focus in Section \ref{sec:setup}, which aims to detect privacy information in real-name LLM interactions.
We also observe that the results of direct prompting are limited to a few common types of privacy, such as gender, with less accuracy in detecting more nuanced aspects like user status or intent. 

To address these limitations, we developed the privacy annotation pipeline, as shown in Figure \ref{fig:framework}. (1) To avoid privacy extraction for user dialogue that does not involve privacy leakage, the first step is a filter, classifying dialogues into two categories: leaked and non-leaked. (2) Since the diverse and open-ended nature of privacy categories in user interactions with LLMs, the second step aims to automatically collect a comprehensive coverage of privacy categories for the corpus. We independently extract privacy categories for each sample and take the union of all categories as the extensive set of privacy categories at the dataset level. (3) Then, with the assistance of the extensive category set, we extract privacy phrases for each sample based on the whole context of queries and filter the results based on the rules of being user-related and having a clear reference. (4) The last step is to annotate privacy information for each extracted phrase.  
We also compare the performance of direct prompting with our automated pipeline in Section \ref{sec:human_eval}. All associated prompts can be found in Appendix \ref{sec:prompt_framework}.

\subsection{Pipeline Design}
\textbf{Privacy Leakage or Not Classification.} 
We explicitly address the scenario where the user interacts with identifier exposure, and prompt GPT-4o to assess whether the user has disclosed any new privacy information during the interaction. We further clarify that privacy information must be linked to the user or individuals associated with them, and it encompasses various aspects such as opinions, preferences, and personal details. To ensure more reliable inferences, we not only ask GPT-4o to generate a simple yes or no judgment, but also request it to explain the reasoning before its decision. Additionally, we provide a set of manually annotated samples as a reference to guide the model.

\textbf{Extensive Privacy Categories Extraction.}\label{sec:category}
We first prompt GPT-4o to directly identify privacy phrases and the corresponding categories for each sample labeled as leaking privacy. We also provide an example to guide GPT-4o. These results are then merged to create an extensive set of privacy categories. 
However, since extraction for each sample is independent, some categories may be expressed differently but hold the same meaning. For example, we may encounter both ``hobby'' and ``interests and hobbies,'' which should be treated as equivalent. To address this, we use GPT-4o to identify such duplicates and assist in the merging process.

Specifically, as shown in Figure \ref{fig:framework}, we partition the shuffled privacy categories into smaller, more manageable blocks and deduplicate categories within each block independently. The deduplicated results from all blocks are then merged. This process is repeated iteratively until no further deduplication occurs, meaning the number of categories remains unchanged before and after the deduplication step. To reduce randomness, we run the algorithm twice and perform a final deduplication by prompting GPT-4o to deduplicate across all categories at once.
In practice, given a seen or even unseen dialogue dataset, performing fine-grained analysis on each sample and then merging ensures comprehensive category coverage at the dataset level and highlights the pipeline’s adaptability and generalizability across diverse dialogue domains.
We also find that using only one-tenth of the data for category extraction yields results that are nearly saturated in terms of diversity. 

\textbf{Privacy Phrase Extraction.}
For each sample that contains privacy leakage, based on the expensive set of privacy categories, we prompt GPT-4o to extract phrases that explicitly reveal privacy information about the user who poses the query. By inserting categories into the prompts, we ask GPT-4o to match the phrases and the categories. The extracted phrase must be directly copied from the query, and the relationship between the phrase and the user should be evident, not speculative. For example, phrases should not be extracted with the reason like ``the user may be interested in something.'' Additionally, any phrases falling outside the given privacy categories, which may indicate a wrong extraction, should be excluded.
To enhance the precision of the extracted phrases, we apply two filtering rules: (1) the privacy phrase must be directly associated with the user or their close associates, and (2) the phrase must have an explicit reference, rather than be a general phrase such as ``go a place'' or ``have ideas''. We use GPT-4o to evaluate these criteria and remove any phrases that do not meet the rules. The effectiveness of filtering based on the rules is shown in Table \ref{table:phrase_eval}.

In practice, samples misclassified as leaking privacy will be corrected if no phrases are extracted. Furthermore, including too many categories in a single input can reduce both precision and recall of phrases. To mitigate this, we divide the categories into smaller blocks and make multiple requests to GPT-4o for each sample, then merge the results. This approach may result in the extraction of synonymous phrases across different requests, such as ``plan to play football'' and ``play football.'' Therefore, we also prompt GPT-4o for deduplication, retaining the more concise and clearer phrase.

\textbf{Privacy Information Annotation.}
Finally, for each extracted phrase, we prompt GPT-4o to annotate the specific privacy information it reveals, based on the context of the whole text. We also provide several manually constructed examples for annotation.

\subsection{Pipeline over Dialogue Datasets}
Based on the automated annotation pipeline, we create large datasets comprising over 249K queries from 33K dialogues with 154K privacy phrases and corresponding information in different languages. 

\textbf{Raw Datasets.} 
We use a diverse set of source corpora, including one English dialogue dataset, ShareGPT \cite{sharegpt}, and three Chinese dialogue datasets, CrossWOZ \cite{crosswoz}, DuConv \cite{duconv}, and LCCC-base \cite{lccc}. The large scale and varied scope of these datasets make them well-suited as our data source. Specifically, ShareGPT naturally consists of human-to-machine interactions, while the three Chinese datasets consist of human-to-human dialogues. To adapt the human-to-human dialogues for our task, we designate the first speaker as the target user for privacy information extraction and the second speaker as the assistant. A detailed introduction of the source corpora is provided in Appendix \ref{sec:intro_source}.

\begin{figure}[!t]
  \centering
         \centering
         \begin{subfigure}{0.66\linewidth}
             \centering
            \begin{minipage}{\linewidth}
             \includegraphics[width=\textwidth]{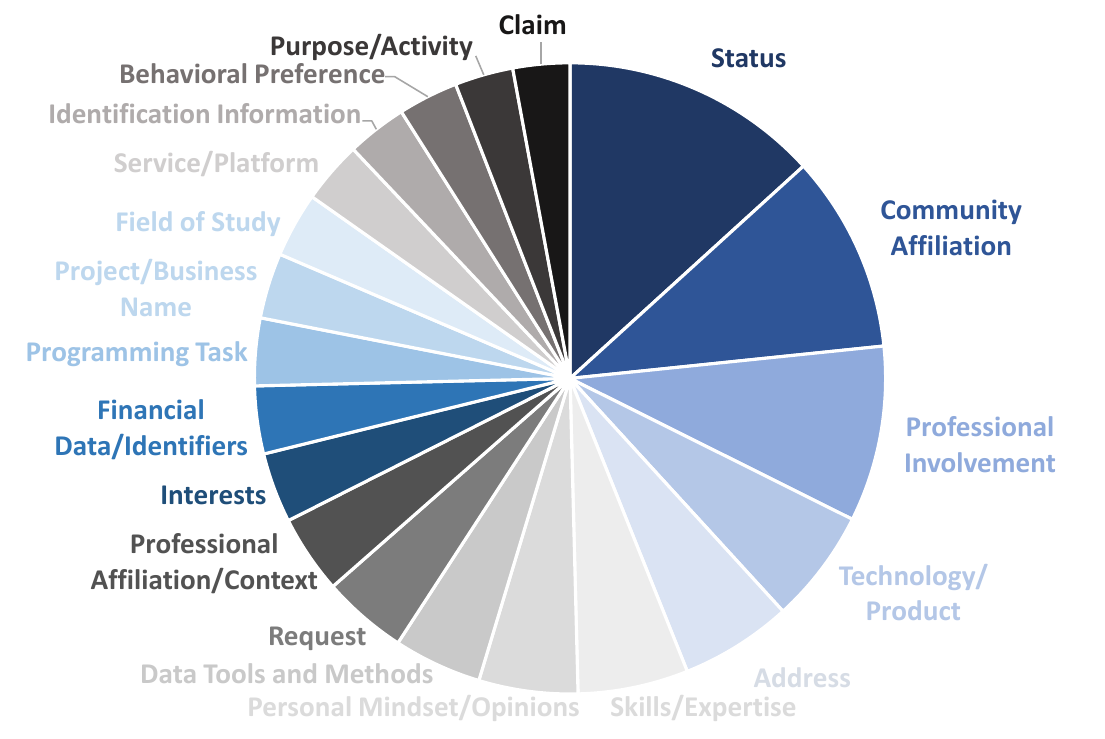}
            \end{minipage}
             \caption{Top 20 privacy categories.}
             \label{fig:category}
        \end{subfigure}
        \begin{subfigure}{0.33\linewidth}
             \centering
            \begin{minipage}{\linewidth}
             \includegraphics[width=\textwidth]{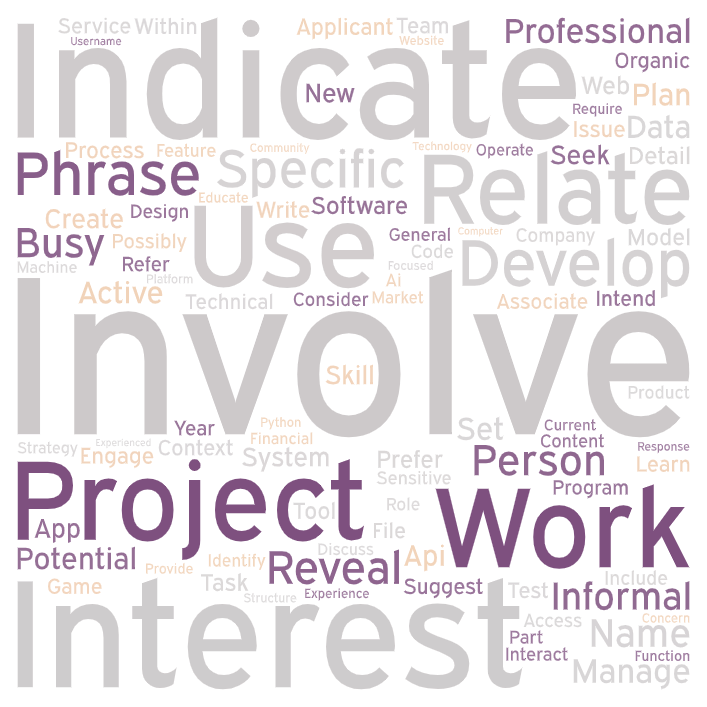}
            \end{minipage}
             \caption{Word cloud.} 
             \label{fig:word_cloud}
        \end{subfigure}
    \caption{Analyses of privacy information in ShareGPT \cite{sharegpt}.}\label{fig:analysis_information} 
\end{figure}

\begin{figure}[!t]
  \centering
         \centering
         \begin{subfigure}{0.66\linewidth}
             \centering
            \begin{minipage}{\linewidth}
             \includegraphics[width=\textwidth]{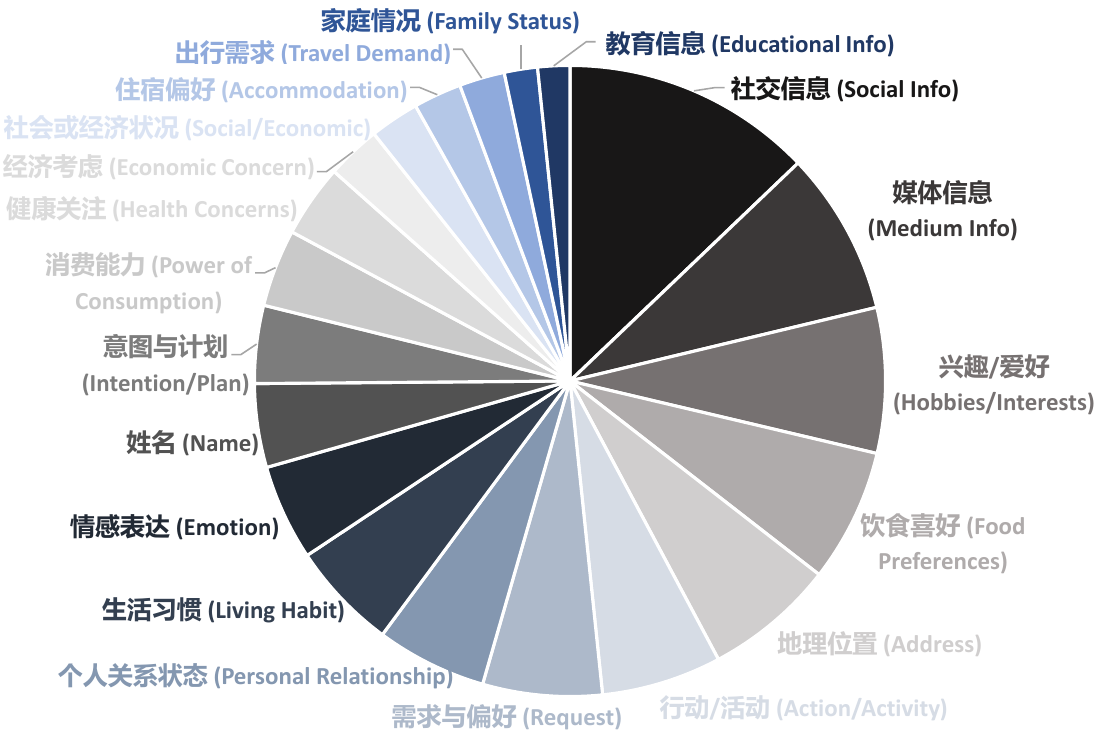}
            \end{minipage}
             \caption{Top 20 privacy categories.}
             \label{fig:category_zh}
        \end{subfigure}
        \begin{subfigure}{0.33\linewidth}
             \centering
            \begin{minipage}{\linewidth}
             \includegraphics[width=\textwidth]{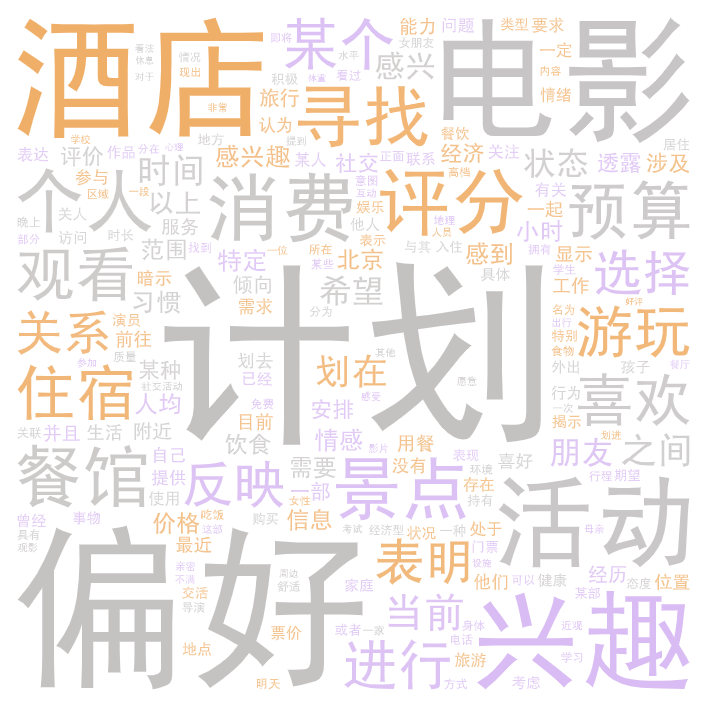}
            \end{minipage}
             \caption{Word cloud.} 
             \label{fig:word_cloud_zh}
        \end{subfigure}
    \caption{Analyses of privacy information in CrossWOZ \cite{crosswoz}, DuConv \cite{duconv}, and LCCC-base \cite{lccc}.}\label{fig:analysis_information_zh}
\end{figure}

\textbf{Annotated Datasets.}
After each step of the automated annotation pipeline, we have: 
(1) classifying privacy leakage obtains over 41K samples with privacy leakage and 56K without privacy leakage in the English portion of the  dataset, and 64K samples with privacy leakage and 88K without privacy leakage in the Chinese portion; 
(2) extracting privacy categories from 2,000 English samples and 5,000 Chinese samples results in 325 categories for English and 149 for Chinese; 
(3) extracting privacy phrases and annotating information obtain 97,659 English samples and 151,988 Chinese samples, of which 32,814 English samples and 43,255 Chinese samples contain privacy, corresponding to 85,320 privacy phrases and information in English and 68,910 in Chinese. Appendix \ref{sec:example_in_out} shows a data example, and the LLM API usage is presented in Appendix \ref{sec:api_cost}.

\textbf{Dataset Analysis.}
Figure \ref{fig:category} and Figure \ref{fig:category_zh} show the relative frequency of the top 20 different privacy categories in the English and the Chinese datasets, respectively. We can find that in open-domain interactions, information such as users' status, opinions, and preferences is more likely to be inadvertently revealed, whereas categories like age and gender appear less frequently. 
Figure \ref{fig:word_cloud} presents a word cloud of the identified privacy information in the English dataset. We can observe that the word cloud reveals a high frequency of terms associated with the user's ``involvement'' in various activities, as well as their personal interests. These results indicate that the privacy information in user interactions with LLM is significantly different from PII in specific documents or content on the web. 
Additionally, a comparison of the word clouds across languages in Figure \ref{fig:word_cloud} and Figure \ref{fig:word_cloud_zh} reveals scenario differences between the English and the Chinese datasets. While the English dataset predominantly features discussions centered on work or project-related topics, the Chinese datasets contain more casual, everyday conversations, often focusing on topics such as films, travel, and leisure activities.

\begin{table}
\centering
\caption{Human evaluation results of phrase extraction for directly prompting GPT-4o vs. our pipeline.}\label{table:phrase_eval}
\resizebox{0.95\linewidth}{!}{
    \begin{tabular}{l|c|c|c|l}
    \toprule
        Method                & $\text{R}_{\mathcal{P}}$ &  $\text{P}_{\mathcal{P}}$ & $\text{F1}_{\mathcal{P}}$ & $\Delta \text{F1}_{\mathcal{P}}$\\
        \midrule
        Directly prompting GPT-4o    & 65.09 & 87.92 & 74.80 & +00.00\\
        \midrule
        Phrase extraction in pipeline & \textbf{97.21} & 85.52 & 90.99 & +16.19\\
        \ \ \ \ + \textit{Filtering rules} & 95.13 & \textbf{92.32} & \textbf{93.70} & \textbf{+18.90}\\
    \bottomrule
    \end{tabular}
    }
\end{table}

\textbf{Comparison with Human Annotation.}\label{sec:human_eval}
To evaluate the privacy leakage classification step, we have 8 annotators classify 400 samples with half in Chinese and half in English, labeling them as either leak or non-leak. 
The guidelines to ensure the objectivity and consistency of the annotation results are provided in Appendix \ref{sec:annotation_guide}. 
The results show an accuracy of 0.96, a recall score of 0.97, and a precision score of 0.94. These metrics demonstrate that the results of the first step closely align with the manual annotations.

We also conduct a human evaluation for the privacy phrase extraction step, as shown in Table \ref{table:phrase_eval}. Specifically, we manually extract privacy phrases for 200 samples to serve as the ground truth, resulting in 351 privacy phrases. Using the evaluation method described in Section \ref{sec:metric}, we found that the recall score for phrases directly extracted by GPT-4o with prompts \cite{ghost} was only 65.09\%, meaning that roughly one-third of privacy phrases were missed.
In comparison to this baseline, our pipeline, by incorporating category information as additional input, shows a slight decrease in precision but significantly increases the number of extracted phrases, boosting the recall score by 32.12\%. As a result, the F1 score improves by 16.19\%. Furthermore, applying filtering rules further enhances the F1 score by an additional 2.71\%. These results also suggest that the phrase extraction step aligns well with manual annotations, highlighting its effectiveness and consistency. 

To further evaluate the quality of the annotated privacy information, we randomly select 200 samples that contain 427 pieces of private information and ask annotators to evaluate the privacy information. Each information is scored as 1 if the generated privacy information accurately reflects the private details of the phrase and is semantically coherent; otherwise, it receives a score of 0. The average final score is 0.93, indicating high-quality and reliable annotation of the privacy information.

\begin{table*}
\begin{minipage}{0.355\textwidth}
\centering
\caption{Accuracy (\%) of different baselines for privacy leakage classification.}\label{table:result_query}
\resizebox{0.90\columnwidth}{!}{
    \begin{tabular}{llrr}
    \toprule
        \multirow{2}{*}{Method} & \makecell[c]{\multirow{2}{*}{Model}} & \multicolumn{1}{c}{English} & \multicolumn{1}{c}{Chinese} \\
        \cmidrule{3-4}
         ~ & ~ & \makecell[c]{Acc} & \makecell[c]{Acc} \\
         \midrule
\multirow{6}{*}{ZG \cite{gpt3}}		&	Llama-3.2-1B	&	$61.18_{\pm 0.49}^*$	&	$58.08_{\pm 0.49}^*$	\\
~	&	Llama-3.2-3B	&	$65.07_{\pm 0.48}^*$	&	$66.97_{\pm 0.47}$	\\
\cmidrule{2-4}
~	&	Qwen2.5-1.5B	&	$59.37_{\pm 0.49}^*$	&	$57.66_{\pm 0.49}^*$	\\
~	&	Qwen2.5-3B	&	$65.77_{\pm 0.47}^*$	&	$65.25_{\pm 0.48}^*$	\\
~	&	Qwen2.5-7B	&	$68.98_{\pm 0.46}^*$	&	$69.67_{\pm 0.46}^*$	\\
~	&	Qwen2.5-72B	&	$73.45_{\pm 0.44}$	&	$75.90_{\pm 0.43}$	\\
\midrule
\multirow{6}{*}{ICL \cite{icl}}	&	Llama-3.2-1B	&	$64.01_{\pm 0.48}^*$	&	$58.32_{\pm 0.49}^*$	\\
~	&	Llama-3.2-3B	&	$60.24_{\pm 0.49}^*$	&	$61.61_{\pm 0.49}^*$	\\
\cmidrule{2-4}
~	&	Qwen2.5-1.5B	&	$63.60_{\pm 0.48}^*$	&	$63.49_{\pm 0.48}^*$	\\
~	&	Qwen2.5-3B	&	$66.42_{\pm 0.47}^*$	&	$61.56_{\pm 0.49}^*$	\\
~	&	Qwen2.5-7B	&	$68.28_{\pm 0.47}^*$	&	$72.21_{\pm 0.45}^*$	\\
~	&	Qwen2.5-72B	&	$71.17_{\pm 0.45}$	&	$76.83_{\pm 0.42}$	\\
\midrule
\multirow{5}{*}{SFT}	&	Llama-3.2-1B	&	$78.69_{\pm 0.41}$	&	$80.99_{\pm 0.39}$	\\
~	&	Llama-3.2-3B	&	$81.63_{\pm 0.39}$	&	$81.32_{\pm 0.39}$	\\
\cmidrule{2-4}
~	&	Qwen2.5-1.5B	&	$82.46_{\pm 0.38}$	&	$86.47_{\pm 0.34}$	\\
~	&	Qwen2.5-3B	&	$82.59_{\pm 0.38}$	&	$87.49_{\pm 0.33}$	\\
~	&	Qwen2.5-7B	&	$\textbf{83.57}_{\pm 0.37}$	&	$\textbf{87.56}_{\pm 0.33}$	\\

        \bottomrule
    \end{tabular}
    }
\end{minipage}
\hfill
\begin{minipage}{0.637\textwidth}
    \centering
    \caption{Performance of different privacy phrase extraction baselines from phrase-level metrics (\%).}\label{table:result_phrase}
\resizebox{0.90\columnwidth}{!}{
    \begin{tabular}{llrrrrrr}
    \toprule
        \multirow{2}{*}{Method} & \makecell[c]{\multirow{2}{*}{Model}} & \multicolumn{3}{c}{English} & \multicolumn{3}{c}{Chinese} \\
        \cmidrule{3-8}
         ~ & ~ & \makecell[c]{$\text{R}_{\mathcal{P}}$} & \makecell[c]{$\text{P}_{\mathcal{P}}$} & \makecell[c]{$\text{F1}_{\mathcal{P}}$} & \makecell[c]{$\text{R}_{\mathcal{P}}$} & \makecell[c]{$\text{P}_{\mathcal{P}}$} & \makecell[c]{$\text{F1}_{\mathcal{P}}$}\\ 
        \midrule
\multirow{6}{*}{ZG \cite{gpt3}}	&	Llama-3.2-1B	&	$29.67_{\pm 0.43}^*$	&	$27.21_{\pm 0.38}^*$	&	28.39	&	$22.07_{\pm 0.40}^*$	&	$26.08_{\pm 0.43}^*$	&	23.91	\\
~	&	Llama-3.2-3B	&	$37.51_{\pm 0.42}^*$	&	$27.86_{\pm 0.42}^*$	&	31.97	&	$52.35_{\pm 0.44}^*$	&	$26.20_{\pm 0.43}^*$	&	34.92	\\
\cmidrule{2-8}
~	&	Qwen2.5-1.5B	&	$36.36_{\pm 0.42}^*$	&	$29.85_{\pm 0.43}^*$	&	32.79	&	$26.73_{\pm 0.43}^*$	&	$25.47_{\pm 0.41}^*$	&	26.08	\\
~	&	Qwen2.5-3B	&	$24.51_{\pm 0.38}^*$	&	$34.09_{\pm 0.45}^*$	&	28.52	&	$20.46_{\pm 0.38}^*$	&	$35.07_{\pm 0.45}^*$	&	25.84	\\
~	&	Qwen2.5-7B	&	$42.83_{\pm 0.43}^*$	&	$42.27_{\pm 0.46}^*$	&	42.55	&	$42.65_{\pm 0.46}^*$	&	$44.74_{\pm 0.48}^*$	&	43.67	\\
~	&	Qwen2.5-72B	&	$61.81_{\pm 0.44}$	&	$44.00_{\pm 0.43}$	&	51.41	&	$72.37_{\pm 0.41}$	&	$50.14_{\pm 0.46}$	&	59.24	\\
\midrule
\multirow{6}{*}{ICL \cite{icl}}	&	Llama-3.2-1B	&	$10.04_{\pm 0.28}^*$	&	$23.93_{\pm 0.39}^*$	&	14.15	&	$20.37_{\pm 0.38}^*$	&	$23.50_{\pm 0.40}^*$	&	21.83	\\
~	&	Llama-3.2-3B	&	$44.04_{\pm 0.42}^*$	&	$28.73_{\pm 0.43}^*$	&	34.77	&	$60.26_{\pm 0.45}^*$	&	$27.96_{\pm 0.42}^*$	&	38.20	\\
\cmidrule{2-8}
~	&	Qwen2.5-1.5B	&	$38.85_{\pm 0.42}^*$	&	$32.78_{\pm 0.44}^*$	&	35.56	&	$39.81_{\pm 0.47}^*$	&	$25.99_{\pm 0.40}^*$	&	31.45	\\
~	&	Qwen2.5-3B	&	$51.37_{\pm 0.41}^*$	&	$30.91_{\pm 0.43}^*$	&	38.60	&	$43.38_{\pm 0.47}^*$	&	$33.56_{\pm 0.44}^*$	&	37.84	\\
~	&	Qwen2.5-7B	&	$55.77_{\pm 0.43}^*$	&	$39.45_{\pm 0.45}^*$	&	46.21	&	$45.36_{\pm 0.47}^*$	&	$49.82_{\pm 0.47}^*$	&	47.49	\\
~	&	Qwen2.5-72B	&	$62.58_{\pm 0.44}$	&	$43.66_{\pm 0.44}$	&	51.43	&	$74.95_{\pm 0.40}$	&	$51.45_{\pm 0.46}$	&	61.02	\\
\midrule
\multirow{5}{*}{SFT}	&	Llama-3.2-1B	&	$65.29_{\pm 0.39}$	&	$52.92_{\pm 0.47}$	&	58.46	&	$71.58_{\pm 0.39}$	&	$62.24_{\pm 0.46}$	&	66.59	\\
~	&	Llama-3.2-3B	&	$67.00_{\pm 0.39}$	&	$56.57_{\pm 0.45}$	&	61.35	&	$75.72_{\pm 0.37}$	&	$64.64_{\pm 0.45}$	&	69.74	\\
\cmidrule{2-8}
~	&	Qwen2.5-1.5B	&	$63.81_{\pm 0.40}$	&	$57.25_{\pm 0.45}$	&	60.35	&	$76.35_{\pm 0.37}$	&	$69.85_{\pm 0.44}$	&	72.96	\\
~	&	Qwen2.5-3B	&	$68.77_{\pm 0.38}$	&	$58.42_{\pm 0.45}$	&	63.18	&	$76.89_{\pm 0.37}$	&	$\textbf{70.74}_{\pm 0.43}$	&	73.69	\\
~	&	Qwen2.5-7B	&	$\textbf{70.75}_{\pm 0.39}$	&	$\textbf{60.31}_{\pm 0.44}$	&	\textbf{65.11}	&	$\textbf{80.15}_{\pm 0.34}$	&	$69.20_{\pm 0.44}$	&	\textbf{74.27}	\\
        \bottomrule
    \end{tabular}
    }
\end{minipage}
\begin{tablenotes}
    \small
    \item ``$*$'' denotes the result is significantly worse than results of corresponding SFT model in t-test with $p < 0.05$.
\end{tablenotes}
\end{table*}

\subsection{Evaluation Metrics}\label{sec:metric}
Dialogue datasets with privacy annotation provide ground truth of privacy leakage, privacy phrases, and corresponding information for local privacy detection methods. We design a hierarchical set of metrics spanning three levels, each aligned with tasks of increasing granularity in privacy detection.

\textbf{Privacy Query-Level Metric for Leakage Classification Task.} 
Given a user's query, the objective of the privacy leakage classification task is to determine whether the query reveals any privacy of the user or user-related individuals. The automatically annotated data can be divided into either ``leaked'' or ``non-leaked,'' based on whether there are privacy phrases extracted. For such query-level classification task, we adopt accuracy to measure the performance of different local privacy detection methods.

\textbf{Privacy Phrase-Level Metrics for Phrase Extraction Task.}
Given a private query, the objective of the privacy phrase extraction task is to extract phrases based on the context of the query that result in the user's privacy leakage.
For each data sample, we focus on both missed and incorrect extractions. For missed extractions, we define the recall score of sample $q$ as the ratio of the intersection between the predicted phrases and the ground truth phrases, normalized by the total number of ground truth phrases. For incorrect extractions, we define the precision score of $q$ as the ratio of the intersection between the predicted phrases and the ground truth phrases, normalized by the total number of predicted phrases. The overall recall and precision scores are calculated as the average of these scores across all samples.

Specifically, we let $\mathcal{Q}_r$ denote the set of user queries where the extracted phrase is non-empty in the ground truth and $\mathcal{Q}_p$ denote the set of user queries where the extracted phrase is non-empty in the prediction results. We let $\hat{\mathcal{P}}_q$ and $\mathcal{P}_q$ denote the set of predicted phrases and the ground truth of the query $q$, respectively. Then, the recall score $\text{R}_{\mathcal{P}}$ and precision score $\text{P}_{\mathcal{P}}$ is defined as 
\begin{equation}
\begin{aligned}
     \text{R}_{\mathcal{P}} = \frac{1}{|\mathcal{Q}_r|}\sum_{q\in \mathcal{Q}_r}\frac{|\hat{\mathcal{P}}_q\cap \mathcal{P}_q|}{|\mathcal{P}_q|},\ \  
     \text{P}_{\mathcal{P}} = \frac{1}{|\mathcal{Q}_p|}\sum_{q\in \mathcal{Q}_p}\frac{|\hat{\mathcal{P}}_q\cap \mathcal{P}_q|}{|\hat{\mathcal{P}}_q|},
\end{aligned}
\end{equation}
where $|\cdot|$ is the size of the set. 
In practice, the extracted phrases may refer to the same entity but have different expressions. To address this issue, our phrase matching strategy goes beyond simply checking whether two phrases are identical. It also considers two phrases a match if one is part of the other or if their Rouge-L score exceeds 0.5 \cite{privacyimplications}.

\textbf{Privacy Information-Level Metrics for Information Summarization Task.}
Given a query with privacy phrases, the objective of the privacy information summarization task is to summarize specific information that is leaked from the query.
In general, we use Rouge-L to evaluate the quality of generated privacy information. Similarly, we define the miss and incorrect extractions of the privacy information. We let $\hat{\mathcal{I}}_q$ and $\mathcal{I}_q$ denote the set of predicted information and the ground truth of the query $q$, respectively. For one privacy information ${I}_j\in {\mathcal{I}}_q$, the recall score is defined as
$
    f_{\text{R}}({I}_j, \hat{\mathcal{I}}_q) = \max_{\hat{I}_k \in \hat{\mathcal{I}}_q} \text{\textit{Rouge}}({I}_j, \hat{I}_k),
$
where $\text{\textit{Rouge}}$ is the function that calculates the Rouge-L score for two sentences. The recall score of ${\mathcal{I}}_q$ is defined as the average recall score of each piece of information it contains. For $\hat{I}_j\in \hat{\mathcal{I}}_q$, the precision score is defined as 
$
    f_{\text{P}}(\hat{I}_j, \mathcal{I}_q) = \max_{I_k \in \mathcal{I}_q} \text{\textit{Rouge}}(\hat{I}_j, I_k).
$
The precision score of $\hat{\mathcal{I}}_q$ is defined as the average precision score of each piece of information it contains.
Then we can define the recall score $\text{R}_{\mathcal{I}}$ and the precision score $\text{P}_{\mathcal{I}}$ for all queries as follow,
\begin{equation}
\begin{aligned}
     \text{R}_{\mathcal{I}} = \frac{1}{|\mathcal{Q}_r|}\sum_{q\in \mathcal{Q}_r}\frac{1}{|\mathcal{I}_q|}{\sum_{I_j \in \mathcal{I}_q}f_{\text{R}}({I}_j, \hat{\mathcal{I}}_q)},\\
     \text{P}_{\mathcal{I}} = \frac{1}{|\mathcal{Q}_p|}\sum_{q\in \mathcal{Q}_p}\frac{1}{|\hat{\mathcal{I}}_q|}{\sum_{\hat{I}_j \in \hat{\mathcal{I}}_q}f_{\text{P}}(\hat{I}_j, {\mathcal{I}}_q)}.
\end{aligned}
\end{equation}
Intuitively, $\text{R}_{\mathcal{I}}$ measures the completeness of the extracted privacy information, while $\text{P}_{\mathcal{I}}$ evaluates the accuracy of the information. 

After getting the recall and precision score, we can also define the F1 score for both phrase-level and information-level as follows,
\begin{equation}
    \text{F1}_{\mathcal{P}} = \frac{2\text{R}_{\mathcal{P}}\text{P}_{\mathcal{P}}}{\text{R}_{\mathcal{P}}+\text{P}_{\mathcal{P}}},\ \ 
    \text{F1}_{\mathcal{I}} = \frac{2\text{R}_{\mathcal{I}}\text{P}_{\mathcal{I}}}{\text{R}_{\mathcal{I}}+\text{P}_{\mathcal{I}}}.
\end{equation}

\section{Evaluation for Local Detection Methods}
In this section, we evaluate various local privacy detection approaches based on the automatically annotated datasets.

\subsection{Setups}
\noindent
\textbf{Dataset Splitting.}
We split the English data at an 8:2 ratio to obtain the training and test set and follow the dataset splits from the original data sources for Chinese data. The detailed training and test set statistics are shown in Appendix \ref{sec:dataset_stat}. 
During the training process, we balance the number of samples with and without privacy leakage in the training data, ensuring a ratio of approximately 1:1. 

\textbf{Models.} 
We take the instruction-tuned version of different series of language models with various scales, including Llama-3.2-1B and 3B \cite{llama3}, and Qwen-2.5-1.5B, 3B, 7B, and 72B \cite{qwen2}. All the pretrained checkpoints are loaded from huggingface\footnote{https://huggingface.co}.

\textbf{Implementation Details.} 
We implement all the baselines in PyTorch. The workstation has 8 NVIDIA V100 32G GPUs. To improve training efficiency, we adopt LoRA \cite{lora} for tuning our model. More training settings are listed in Appendix \ref{training_detail}.

\begin{figure*}[!t]
  \centering
    \includegraphics[width=0.92\textwidth]{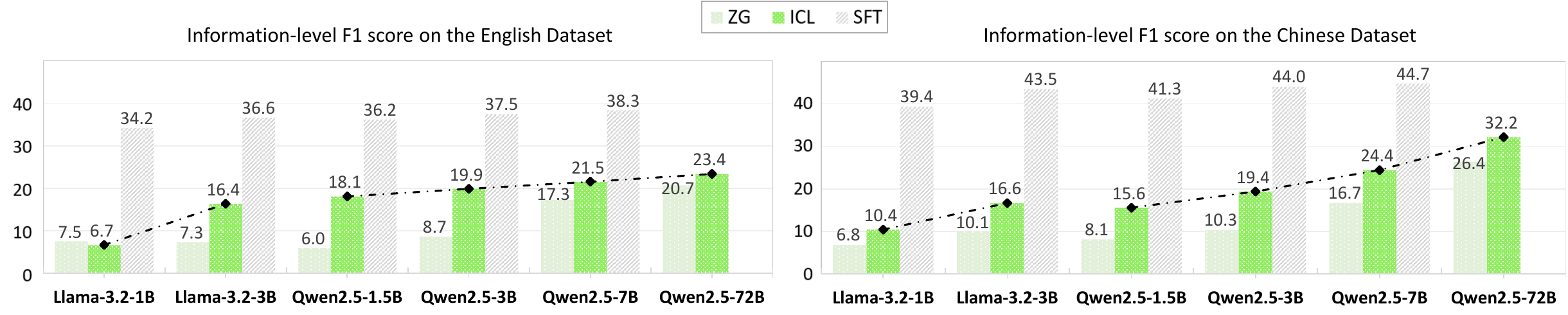}
    \caption{Performance of different privacy information summarization baselines from information-level F1 (\%).}
  \label{fig:result_info}
\end{figure*}

\subsection{Local Privacy Detection Baselines}

\textbf{Tuning-free Baselines}
Tuning-free methods leverage the inherent capabilities of pre-trained models for privacy detection.  
(1) \textbf{Zero-shot Generation (ZG)} \cite{gpt3}.
We directly prompt local LLMs to determine whether the query leaks privacy or to extract privacy phrases and corresponding information. In the prompt, we explicitly specify that privacy information in user dialogue includes various aspects such as opinions, preferences, intentions, etc. We also require that the recognized privacy be relevant to the user and ask the model to think step by step. Finally, we guide the model to output structured results in the specified format. 
(2) \textbf{In-Context Learning (ICL)} \cite{icl}.
Based on the prompts of ZG, we further insert five examples randomly sampled from training data based on the template for inference with local privacy detection LLM. After the examples, the current user query is added for privacy detection. All corresponding prompts are provided in Appendix \ref{sec:eval_prompt}.

\textbf{Tuning-based Baseline.}
We use \textbf{Supervised Fine-Tuning (SFT)} with instructions to enable the model to judge privacy leakage or to generate structured phrases and corresponding information based on the given conversations.
Additionally, we find that overly long instruction templates will reduce the model's sensitivity to user queries, resulting in a performance drop. The instruction template used during training is also provided in Appendix \ref{sec:eval_prompt}.

\subsection{Main Results}

\textbf{Privacy Leakage Classification.}
We first evaluate the performance of different baselines on the binary classification task, and the results are presented in Table \ref{table:result_query}. We can find that model performance generally improves with an increase in parameter size. Notably, the largest model in our evaluation, Qwen2.5-72B, excels with ZG, achieving 14.08\% and 18.24\% improvement for query-level accuracy compared to Qwen2.5-1.5B in English and Chinese datasets, respectively. 
By comparing SFT with ZG and ICL across different models, we can find that the fine-tuned Qwen2.5-7B achieves the best performance on both English and Chinese datasets. Specifically, SFT with Qwen2.5-7B outperforms ZG and ICL with the same model by over 14.59\% and 15.35\%, respectively.

\textbf{Privacy Phrase Extraction.} 
We then investigate the quality of extracted privacy phrases by different baselines. From Table \ref{table:result_phrase}, we can also observe that ICL outperforms ZG in most cases. For example, with the Qwen2.5-7B, ICL achieves 3.66\% and 3.82\% improvement for phrase-level F1 over ZG with the same model in English and Chinese datasets, respectively. These results suggest that inserting examples from the training set enhances privacy detection. Additionally, by fine-tuning on our dataset, the 1B model outperforms the direct inference results of the 72B model by 7.03\% and 5.57\% in English and Chinese datasets, respectively, which is friendly for local devices. Furthermore, the fine-tuned Qwen2.5-7B achieves an improvement of 13.68\% and 13.25\% for phrase-level F1 compared to Qwen2.5-72B with ICL in English and Chinese datasets, respectively. These significant gains underscore the importance of tuning with our custom dataset. 

\textbf{Privacy Information Summarization.}
Figure \ref{fig:result_info} presents the performance of different local privacy detection baselines from information-level F1. We can find that by supervised fine-tuning with the training set, Qwen2.5-7B achieves the best performance on different datasets. Specifically, the fine-tuned Qwen2.5-7B outperforms ZG with Qwen2.5-7B by 21.06\% and 28.03\%, and outperforms ICL with Qwen2.5-7B by 16.80\% and 20.33\% in English and Chinese datasets, respectively. 
We can also observe that the fine-tuned 1B model outperforms both ZG and ICL with the 72B model by over 10.83\% and 7.18\% in English and Chinese datasets, respectively.

\textbf{Discussion.}
Although the fine-tuned Qwen2.5-7B model achieves a binary privacy leakage classification accuracy of 87.56\%, its performance on more complex tasks remains insufficient for practical deployment. For instance, the best fine-tuned model attains a phrase-level F1 score of 74.27\% and an information-level F1 score of only 44.70\% on the Chinese dataset. These results emphasize the need for exploring more effective methods grounded in our dataset. 
Furthermore, to explore the real-time inference efficiency of the local fine-tuned models, we also measured the average privacy detection time per query in Appendix \ref{sec:infer_time}.

\subsection{Comparison with Traditional PII Detection}\label{sec:traditional_pii}

\begin{table}
    \centering
    \caption{Performance of traditional PII detection methods from phrase-level metrics (\%)}\label{table:result_traditional_pii}
\resizebox{0.92\columnwidth}{!}{
    \begin{tabular}{lrrrrrr}
    \toprule
 \makecell[c]{\multirow{2}{*}{Method}} & \multicolumn{3}{c}{English} & \multicolumn{3}{c}{Chinese} \\
        \cmidrule{2-7}
          ~ & \makecell[c]{$\text{R}_{\mathcal{P}}$} & \makecell[c]{$\text{P}_{\mathcal{P}}$} & \makecell[c]{$\text{F1}_{\mathcal{P}}$} & \makecell[c]{$\text{R}_{\mathcal{P}}$} & \makecell[c]{$\text{P}_{\mathcal{P}}$} & \makecell[c]{$\text{F1}_{\mathcal{P}}$}\\ 
        \midrule
Presidio \cite{presidio}         &   13.51 & 25.15 & 17.58 & 19.38 & 22.64 & 20.89 \\
Presidio [transformer] & 14.73 & 27.13 & 19.09 & 18.52 & 22.68 & 20.39 \\
\midrule
Qwen2.5-7B with SFT    	&	\textbf{70.75}	&	\textbf{60.31}	&	\textbf{65.11}	&	\textbf{80.15}	&	\textbf{69.20}	&	\textbf{74.27}	\\
        \bottomrule
    \end{tabular}
    }
\end{table}

We also evaluate open-source traditional PII detection methods on our datasets, including Presidio with its default spaCy-based model and an alternative configuration using a transformer-based named entity recognition model \cite{presidio}. The results, presented in Table \ref{table:result_traditional_pii}, show that these traditional methods perform substantially worse than the Qwen2.5-7B model fine-tuned on our dataset. This performance gap also indicates that the nature of privacy information in user interactions with LLMs differs markedly from the PII typically found in structured documents or publicly available web content.

\subsection{Case Study}

\begin{figure}[!t]
    \centering
        \includegraphics[width=0.95\linewidth]{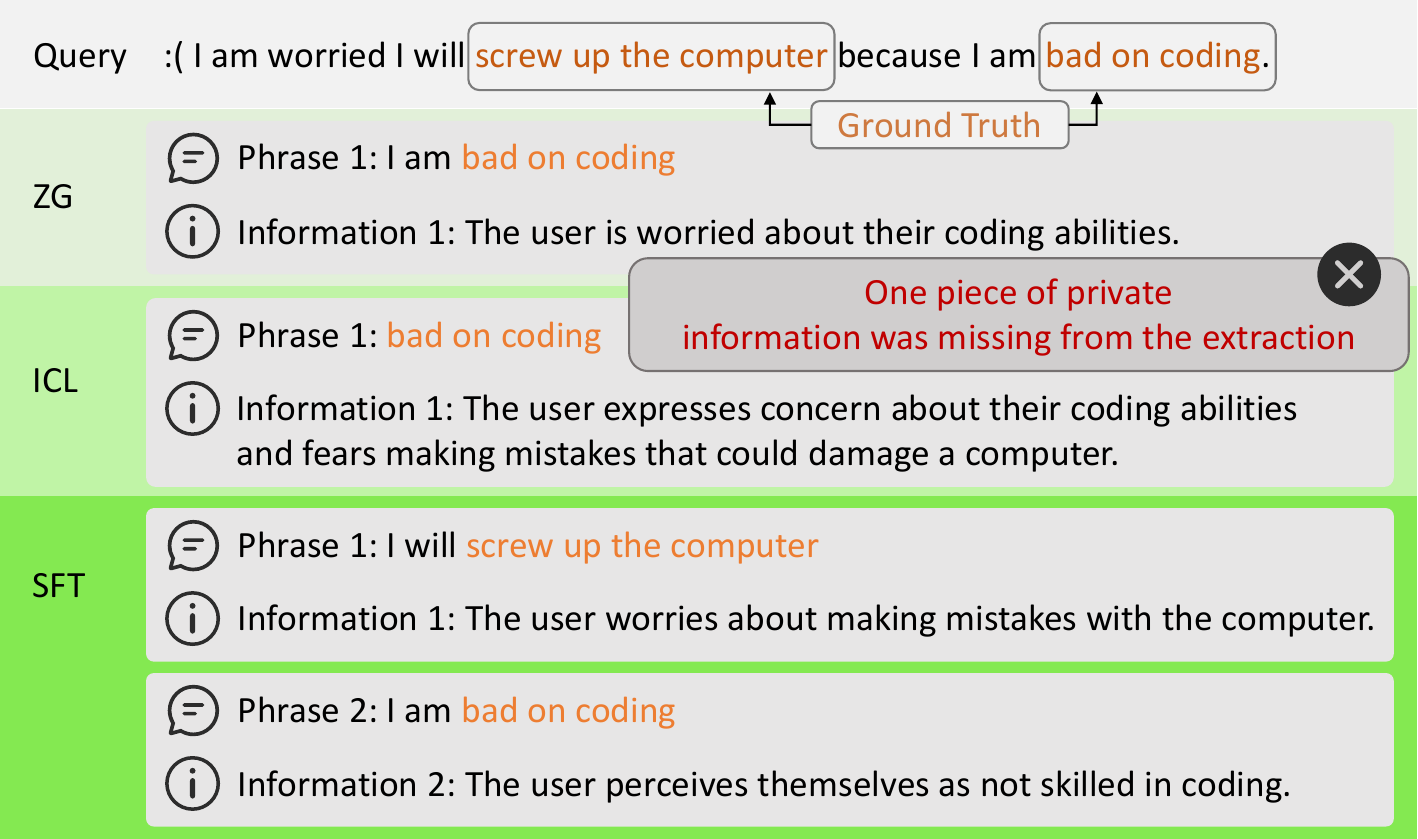}
        \caption{Case study with Qwen2.5-7B-Instruct. SFT successfully detects all relevant privacy details.}\label{fig:case_study}
    \vspace{-0.4em}
\end{figure}

To better assess the performance of various local privacy detection methods, we present a case study of phrase and information extraction in Figure \ref{fig:case_study}. We can observe that both ZG and ICL fail to capture the user's specific concerns, while ICL generates more detailed privacy information compared to ZG. In contrast, SFT successfully identifies all relevant privacy phrases from the user input and generates accurate privacy information. We present a failure case for further analysis of these baselines' performance in Appendix \ref{sec:case_study}.

\section{Conclusion}

In this work, we have studied privacy detection for real-name LLM interaction. We have constructed the first large-scale multilingual dialogue dataset for different privacy detection tasks, achieved through an automated scalable privacy annotation pipeline for privacy phrase extraction and corresponding information annotation. 
Additionally, we have developed evaluation metrics at the query-level, phrase-level, and information-level. 
Finally, we have evaluated various local privacy detection methods, revealing that privacy detection in LLM interactions remains a challenging task. 
As for future research opportunities, the dataset provides a valuable foundation to explore more effective privacy detection methods that can be efficiently deployed on different local devices.

\bibliographystyle{ACM-Reference-Format}
\bibliography{sample-base}

\clearpage
\appendix



\section{Prompts for Automated Privacy Annotation Pipeline}\label{sec:prompt_framework}

Figure \ref{table:prompt_pipeline_step1}, Figure \ref{table:prompt_pipeline_step2} - \ref{table:prompt_pipeline_merge}, Figure \ref{table:prompt_pipeline_step3} - \ref{table:prompt_pipeline_step34}, and Figure \ref{table:prompt_pipeline_step4} provides prompts for privacy leakage classification, extensive privacy categories extraction, privacy phrase extraction, and privacy information annotation in designed pipeline, respectively. 

\section{Additional Dataset Details}
\subsection{Introduction of Data Source}\label{sec:intro_source}
ShareGPT \cite{sharegpt} is composed of dialogues between users and ChatGPT, with 79K raw dialogues and 243K user queries. CrossWOZ \cite{crosswoz} is a task-oriented dialogue dataset covering five domains, containing 6K dialogues and 47K user queries. DuConv \cite{duconv} focuses on movie and film star topics, with 22K dialogues and 99K user queries. Both CrossWOZ and DuConv data are collected through dialogues conducted by two workers, based on specific topics or knowledge backgrounds. The LCCC-base \cite{lccc} is derived from the Weibo corpus. We selected a subset of this dataset as our source data, including 50K dialogues and 76K user queries.

\subsection{Dataset Examples}\label{sec:example_in_out}

Figure \ref{table:dataset_example} shows raw data examples and examples of the constructed dataset. One data sample consists of a user-initiated query and a response from the assistant. Based on this, we add privacy detection results to each query, in the form of a list of privacy. Each element in the list consists of a phrase extracted from the query and the corresponding specific privacy information. In terms of dataset processing, we also retain the contextual relationship between the original queries in one dialogue.

\begin{table}[!t]
\centering
\caption{Statistics of the dataset with automatically annotated privacy phrases and information.}\label{table:dataset_statistics}
\resizebox{0.95\linewidth}{!}{
    \begin{tabular}{lllllll}
    \toprule
        \multirow{2}{*}{Language} & \multicolumn{3}{c}{Training set} & \multicolumn{3}{c}{Test set}\\
        \cmidrule{2-7}
        ~  & \# Non-Leak & \# Leak & \# Phrase & \# Non-Leak & \# Leak & \# Phrase\\
        \midrule
        English   & 52,053 & 26,156 & 67,780 & 12,792 & 6,658 & 17,540  \\
    \midrule
        Chinese   & 98,100 & 38,025 & 60,611 & 10,633 & 5,230 & 8,299\\
    \bottomrule
    \end{tabular}
    }
\end{table}

\subsection{Details on Training and Testing Sets}\label{sec:dataset_stat}

Table \ref{table:dataset_statistics} lists detailed statistics of the dialogue dataset with annotated privacy phrases and information. The privacy phrase set and information set are empty for samples without privacy leakage, while the average length of the phrase set and information set for samples with privacy leakage is 2.6 in the  English part and 1.6 in the Chinese part.

\begin{table*}[!t]
\centering
\caption{Average privacy detection time per query (s) of fine-tuned local models for English samples.}\label{table:infer_time_en}
\resizebox{1.25\columnwidth}{!}{
    \centering
    \begin{tabular}{lrrrrr}
    \toprule
    Local Device	& Llama3.2-1B	&  Qwen2.5-1.5B    &	Qwen2.5-3B    &	Llama3.2-3B   &	Qwen2.5-7B \\
    \midrule
    GeForce RTX 2080 Ti	& 0.1772	& 0.2113	& 0.2668	& 0.4062	& out-of-memory \\
    GeForce RTX 4090	& 0.1191	& 0.1209	& 0.1630	& 0.2025	& 0.3980 \\
    \bottomrule
    \end{tabular}
    }
\end{table*}

\begin{table*}[!t]
\centering
\caption{Average privacy detection time per query (s) of fine-tuned local models for Chinese samples.}\label{table:infer_time_zh}
\resizebox{1.25\columnwidth}{!}{
    \centering
    \begin{tabular}{lrrrrr}
    \toprule
    Local Device	& Llama3.2-1B	&  Qwen2.5-1.5B    &	Qwen2.5-3B    &	Llama3.2-3B   &	Qwen2.5-7B \\
    \midrule
GeForce RTX 2080 Ti	& 0.0885	& 0.0922	& 0.1279	& 0.1623 	& out-of-memory \\
GeForce RTX 4090	& 0.0521	& 0.0623	& 0.0789	& 0.0879 	& 0.1754 \\
    \bottomrule
    \end{tabular}
    }
\end{table*}

\section{Human Annotation and Evaluation Guidelines}\label{sec:annotation_guide}

\subsection{Annotation Instructions for Privacy Leakage}
Annotators are eight students in the lab. They are instructed to determine whether the query reveals privacy information about the user who posed it. Given that the user’s identifier is exposed, annotators must assess whether the query discloses any new privacy information about the user or related individuals or entities. Privacy information includes opinions, concerns, preferences, activities, intentions, or any other information that could be considered private or sensitive based on the query's context. Annotators' responses should only be ``true'' or ``false.''

\subsection{Annotation Instructions for Privacy Phrase Extraction}
Based on the definition of privacy information in user dialogues, annotators are tasked with extracting phrases that reveal private details from the query. These extracted phrases should be directly copied from the query without any alterations to their format or wording. The selected phrases must clearly identify and reflect the user's private information, serving as key indicators of such information. Each annotated result is reviewed for accuracy by a second annotator to ensure consistency and reliability.

\subsection{Evaluation Instructions for Annotated Privacy Information}
Annotators are asked to evaluate the annotated privacy information corresponding to each privacy phrase. A score of 1 is assigned if the generated privacy information accurately captures the details of the phrase and is semantically coherent; otherwise, it receives a score of 0. The final score is computed as the average of all individual scores.

\begin{figure*}[!t]
    \centering
        \includegraphics[width=0.95\linewidth]{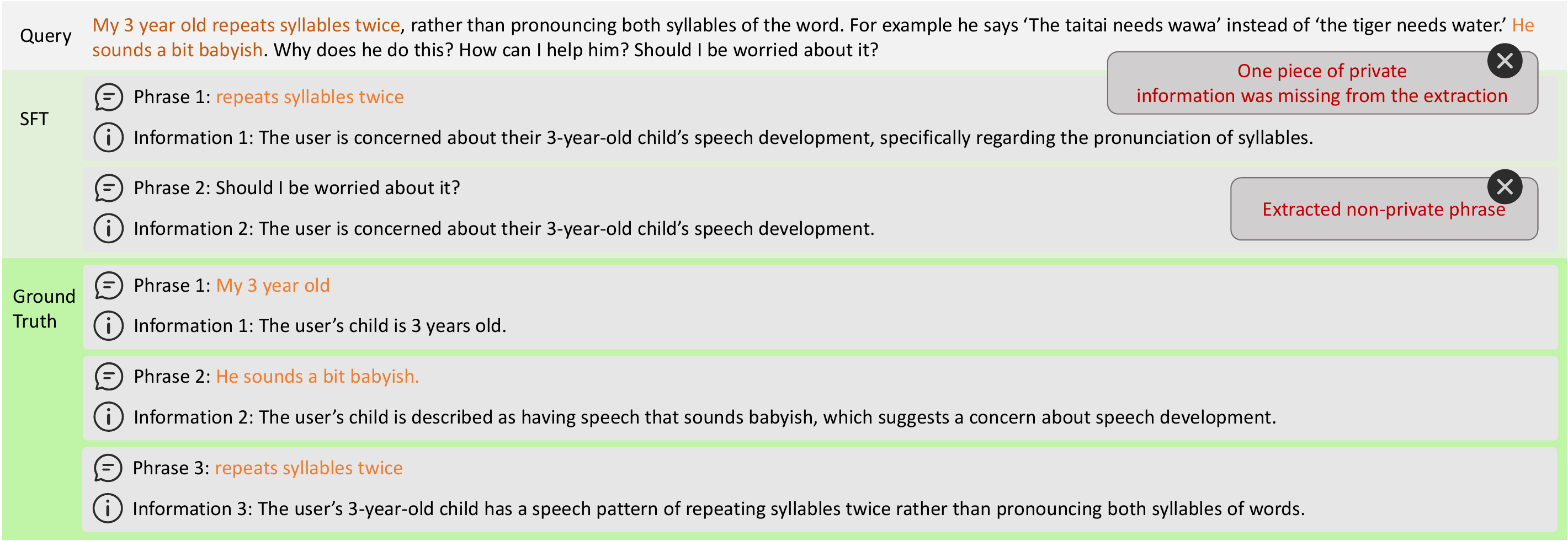}
        \caption{A failure case of Qwen2.5-7B-Instruct with SFT.}\label{fig:bad_case}
\end{figure*}

\section{Additional Experiment Details and Results}
\subsection{Additional Training Details}\label{training_detail}
We utilize the AdamW optimization scheme with PyTorch for all methods. For Llama3.2-1B-Instruct and Qwen2.5-1.5B-Instruct, the learning rate is set between a maximum of 2e-3 and a minimum of 1e-3. For models larger than 1.5B parameters, the learning rate is set to a maximum of 2e-4 and a minimum of 1e-4. Training is conducted with a batch size of 4, for 10 epochs, and a maximum token length of 1024.
For LoRA, we configure the rank to 16, apply a dropout rate of 0.05, and perform weight decomposition on the query, key, and value matrices, as well as the dense MLP layers in the transformer blocks. Additionally, during training, the parameters of Qwen and Llama are set to FP32 to maintain precision.

\subsection{Prompts for Local Privacy Detection Methods}\label{sec:eval_prompt}
Figure \ref{table:prompt_zero_query} - Figure \ref{table:prompt_zero_info}, Figure \ref{table:prompt_icl_query} - \ref{table:prompt_icl_info}, and Figure \ref{table:prompt_train_query} - \ref{table:prompt_train_info} shows detailed prompts used for zero-shot generation, in-context learning, and training privacy detection models, respectively. For zero-shot generation and in-context learning, we prompt language models to extract privacy phrases from user dialogues and generate specific information for each identified phrase. For fine-tuning, we employ a concise template to guide the model, allowing it to learn both the content and output format from the training data.

\subsection{Inference Time for Local Privacy Detection Models}\label{sec:infer_time}
To explore the real-time inference efficiency of the local privacy detection models, which are fine-tuned on our dataset, we measured the average privacy detection time per query for both English and Chinese samples. The model precision is set to FP16, and the results are shown in Table \ref{table:infer_time_en} and Table \ref{table:infer_time_zh}. These results reveal the time efficiency of local privacy detection with billion-scale LLMs.

\subsection{Case Study}\label{sec:case_study}

Figure \ref{fig:bad_case} presents a failure case from the fine-tuned Qwen2.5-7B, where the local model incorrectly extracts a general question that does not actually expose the user's privacy. Moreover, the model fails to identify certain implicit privacy phrases. These shortcomings highlight the challenges of extracting phrases and generating accurate information. Light-weight LLMs, with limited text comprehension abilities, require more sophisticated methods to be developed in future work for effective privacy detection on the user side.

\section{API Cost}\label{sec:api_cost}
We use gpt-4o-2024-08-06 as the base LLM for our automated privacy annotation pipeline, which extracts privacy phrases and annotates information. For the English dataset, the total token count is 296M tokens for input and 152M for output. For the Chinese dataset, the total token count is 219M tokens for input and 99M for output. The total annotation cost amounts to approximately \$3,800, and the average cost is \$0.015 per sample.

\section{Ethical Statement}\label{sec:ethical_statement}
We investigate privacy detection in users' real-name interactions with LLM. The design, implementation, and use of the automated privacy annotation pipeline, along with the local privacy detection methods, are guided by ethical principles for responsible AI deployment.
(1) Dataset Construction: All dialogue data used in this study are publicly available and properly cited, with no ethical concerns regarding their use for research purposes.
(2) Fair Usage: The primary goal of the privacy annotation pipeline is to extract privacy phrases and annotate privacy information from public dialogue datasets, rather than conducting malicious analysis of user privacy. We believe that the responsible application of our design aligns with ethical guidelines. 
Furthermore, to apply our annotation pipeline to a new corpus, such as private enterprise documents, one can begin by annotating a small subset of the preprocessed data with our pipeline. A privacy detection model can then be trained on this subset to generate labels for the remaining data. Note that all sensitive content uploaded to GPT-4o or any cloud-based LLM should be carefully reviewed to ensure data security and compliance.

\begin{figure*}[!t]
  \centering
    \begin{tcolorbox}[title=Prompts for privacy leakage classification (English).]
\footnotesize
A user poses a query\\
"<|INPUT|>"\\
to you. \\
~\\
Before the user poses the query, you only know the \textcolor{flesh}{real unique identifier of the user}, and do not know any other privacy information of the user. \\
After receiving the query with the user's real unique identifier, think step by step to judge whether the query \textcolor{flesh}{reveals any new privacy information of the user or the user's related people or things} to you. \\
~\\
The privacy information should be relevant to the user or the user's related people or things based on the context of the query.\\
The privacy information includes any opinions, concerns, preferences, activities, intentions, or any other information that could be considered private or sensitive based on the context of the query.\\
~\\
Give your think steps. \textcolor{flesh}{Your reasoning must depend on the direct evidence contained in the query and cannot be based on assumption}.\\
Then, output a json as follows:
\begin{lstlisting}[language=plainjson]
```json
{
    "reason": reason for your judgment,
    "judgment": true or false
}
''' 
\end{lstlisting}

Here are some examples:
User's query: We want to play football today.
JSON Output:
\begin{lstlisting}[language=plainjson]
```json
{
    "reason": "This query exposes the activity of the user and people related to the user intend to do.",
    "judgment": true
}
'''
\end{lstlisting}
...... (other examples) 
    \end{tcolorbox}

\begin{tcolorbox}[title=Prompts for privacy leakage classification (Chinese).]
\footnotesize 
\chinese{用户向你提出一个请求}\\
\chinese{"<|INPUT|>"}\\
\chinese{}\\
\chinese{在用户提出请求之前，你只知道\textcolor{flesh}{用户的真实的唯一标识符}，并且不知道用户的任何其他的隐私敏感信息。}\\
\chinese{在收到带有用户真实的唯一标识符的请求后，请你进行逐步的思考（Think Step By Step）来判断该请求是否向你\textcolor{flesh}{透露了任何新的关于用户或与用户相关的人或事的隐私敏感信息}。}\\
\chinese{}\\
\chinese{隐私敏感信息应与用户或与用户相关的人或事相关，该相关性体现在用户请求的上下文中。}\\
\chinese{隐私敏感信息包括任何观点、担忧、偏好、活动、意图或任何其他根据查询上下文可被视为私密或敏感的信息。}\\
\chinese{~}\\
\chinese{给出你的思考与分析步骤。\textcolor{flesh}{你的推理必须依赖于查询中包含的直接证据，不能基于假设}。}\\
\chinese{}\\
\chinese{最后，输出如下格式的json：}
\begin{lstlisting}[language=plainjson]
```json
{
    "reason": (*@\chinese{\footnotesize 你判断的理由,}@*)
    "judgment": true or false
}
'''
\end{lstlisting}
\chinese{以下是一些示例：}\\
\chinese{用户的请求：我们今天想踢足球。}\\
\chinese{JSON输出：}
\begin{lstlisting}[language=plainjson]
```json
{
    "reason": "(*@\chinese{\footnotesize 该请求透露了用户及与用户相关的人打算进行的活动。}@*)",
    "judgment": true
}
'''
\end{lstlisting}
\chinese{...... (other examples)}
\end{tcolorbox}

    \caption{Prompt template for privacy leakage classification for English and Chinese dataset, where <|INPUT|> is the input of the user query.}
    \label{table:prompt_pipeline_step1}
\end{figure*}

\begin{figure*}[!t]
  \centering
    \begin{tcolorbox}[title=Prompts for privacy categories extraction (English).]
\footnotesize
Given a user's query \\
"<|INPUT|>"\\
\\
Identify which \textcolor{flesh}{phrases in the query leak the user's privacy information and the corresponding categories of privacy information}.\\
Let's think step by step and output a json finally.\\
The result output should be in json format with the following format:
\begin{lstlisting}[language=plainjson]
```json
{
    privacy information category 1: [phrase 1, phrase 3, ...], 
    privacy information category 2: [phrase 2, phrase 4, ...]
}
'''
\end{lstlisting}

Here is an example\\
"Query": "I'm a teacher in China, majoring in computer science. I will go back home to London."\\
Output:
\begin{lstlisting}[language=plainjson]
```json
{
    "Job": ["teacher"],
    "Subject": ["computer science"],
    "Location": ["China", "London"]
}
'''
\end{lstlisting} 
\end{tcolorbox}

\begin{tcolorbox}[title=Prompts for privacy categories extraction (Chinese).]
\footnotesize 
\chinese{用户向你提出一个请求}\\
\chinese{"<|INPUT|>"}\\
\chinese{}\\
\chinese{提取请求中\textcolor{flesh}{泄露用户隐私信息的短语，并将其分类为相应的隐私信息类别}。}\\ 
\chinese{让我们一步一步地思考，最后输出一个json，格式如下：}
\begin{lstlisting}[language=plainjson]
```json
{
    (*@\chinese{\footnotesize "隐私信息类别1"}@*): [(*@\chinese{\footnotesize "短语1", "短语3"}@*), ...],
    (*@\chinese{\footnotesize "隐私信息类别2"}@*): [(*@\chinese{\footnotesize "短语2", "短语4"}@*), ...]
}
'''
\end{lstlisting}
\chinese{以下是一个示例：}\\
\chinese{用户的请求："我是中国的一名教师，主修计算机科学。我将回家到伦敦。"}\\ 
\chinese{JSON输出：}
\begin{lstlisting}[language=plainjson]
```json
{
    (*@\chinese{\footnotesize "职业"}@*): [(*@\chinese{\footnotesize "教师"}@*)],
    (*@\chinese{\footnotesize "专业"}@*): [(*@\chinese{\footnotesize "计算机科学"}@*)],
    (*@\chinese{\footnotesize "地点"}@*): [(*@\chinese{\footnotesize "中国", "伦敦"}@*)]
}
'''
\end{lstlisting}
\end{tcolorbox}

    \caption{Prompt template for privacy categories extraction in step 2 for English and Chinese datasets, where <|INPUT|> is the input of the user query.}\label{table:prompt_pipeline_step2}
\end{figure*}

\begin{figure*}[!t]
  \centering
    \begin{tcolorbox}[title=Prompts for categories deduplication (English).]
\footnotesize
Given the following <|BLOCK-SIZE|> privacy categories and the list of example phrases belonging to each privacy category:\\
<|INPUT|>\\
\\
\textcolor{flesh}{If privacy category A and privacy category B have the same meaning, just keep category A}.\\
Let's think step by step and output a json list finally. \\
Output the new privacy categories after deduplication in the format of a list in json, for example:
\begin{lstlisting}[language=plainjson]
[privacy category 1, privacy category 2, ... ]
\end{lstlisting} 
\end{tcolorbox}

\begin{tcolorbox}[title=Prompts for categories deduplication (Chinese).]
\footnotesize 
\chinese{给定以下<BLOCK-SIZE>个隐私类别及每个隐私类别对应的示例短语列表：}\\
\chinese{<|INPUT|>}\\
\chinese{}\\
\chinese{对\textcolor{flesh}{隐私类别进行去重，如果隐私类别A和隐私类别B具有相同的含义，只保留类别A}。}\\
\chinese{让我们一步一步地思考，最后输出一个json列表。}\\
\chinese{以列表格式输出去重后的新的隐私类别，例如：}
\begin{lstlisting}[language=plainjson]
[(*@\chinese{\footnotesize 隐私类别1, 隐私类别2}@*), ... ]
\end{lstlisting} 
\end{tcolorbox}

    \caption{Prompt template for categories deduplication in step 2 for English and Chinese dataset, where <|INPUT|> is the input of privacy categories and corresponding phrase examples, and <|BLOCK-SIZE|> is the number of input categories.}\label{table:prompt_pipeline_merge}
\end{figure*}

\begin{figure*}[!t]
  \centering
    \begin{tcolorbox}[title=Prompts for privacy phrase extraction (English).]
\footnotesize
A user poses a query\\
"<|INPUT|>" to you. \\
\textcolor{flesh}{Extract informative phrases} from the query that definitely reveal the privacy-sensitive information of the user, or user related people or things. \\
The phrases must be included exactly as they appear in the query's content. Do not alter the format or wording of the phrases. For example, spelling errors, number of spaces, capitalization, etc., must all be preserved exactly same as the query's content.\\
Give your think steps. \textcolor{flesh}{The reasoning must depend on the direct evidence contained in the query and cannot be based on assumption}.\\
Then, \textcolor{flesh}{match each extracted phase with the most appropriate privacy category} in the following pre-defined list, and also note if the extracted phrase is not matched with any given privacy category, it should be excluded.\\
Privacy Categories: <|CATEGORIES|>\\
The result output should be in json format with the following format:
\begin{lstlisting}[language=plainjson]
```json
{
    "phrase 1": "privacy category 3",
    "phrase 2": "privacy category 4"
}
'''
\end{lstlisting}
Here are some examples\\
User's Query: "I'm a teacher in China, majoring in computer science. I will go back home to London."\\
Privacy Categories: $\text{[}$["Name", "Age", "Gender", "Job", "Subject", "Location"$\text{]}$ \\
JSON Output:
\begin{lstlisting}[language=plainjson]
```json
{
    "teacher": "Job",
    "computer science": "Subject",
    "China": "Location",
    "London": "Location"
}
'''
\end{lstlisting}
...... (other examples)
\end{tcolorbox}

\begin{tcolorbox}[title=Prompts for privacy phrase extraction (Chinese).]
\footnotesize 
\chinese{用户向你提出一个请求}\\
\chinese{"<|INPUT|>"}\\
\chinese{从请求的内容中\textcolor{flesh}{提取短语}，该短语明确揭示用户的，或与用户相关的人或事物的隐私敏感信息。}\\
\chinese{短语必须与请求内容中出现的完全一致。不要更改短语的格式或措辞，即拼写错误、空格数量、大小写等，都必须与请求内容完全相同。}\\  
\chinese{给出你的思考步骤（Think Step by Step）。\textcolor{flesh}{推理必须依赖于请求中包含的直接证据，不能基于假设}。}\\
\chinese{然后，\textcolor{flesh}{将每个提取的短语与以下预定义列表中最合适的隐私类别匹配}，并且如果提取的短语未与任何给定的隐私类别匹配，则应将其排除。}\\
\chinese{}\\
\chinese{隐私类别：<|CATEGORIES|>}\\
\chinese{结果输出应为json格式，格式如下：}
\begin{lstlisting}[language=plainjson]
```json
{
    (*@\chinese{\footnotesize "短语1": "隐私类别3",}@*)
    (*@\chinese{\footnotesize "短语2": "隐私类别4"}@*)
}
'''
\end{lstlisting}
\chinese{以下是一些示例}\\
\chinese{用户的请求："我是中国的一名教师，主修计算机科学。我将回家到伦敦。"}\\
\chinese{隐私类别：$\text{[}$"姓名", "年龄", "性别", "职业", "专业", "地点"$\text{]}$}\\
\chinese{JSON输出：}
\begin{lstlisting}[language=plainjson]
```json
{
    (*@\chinese{\footnotesize "教师": "职业",}@*)
    (*@\chinese{\footnotesize "计算机科学": "专业",}@*)
    (*@\chinese{\footnotesize "中国": "地点",}@*)
    (*@\chinese{\footnotesize "伦敦": "地点"}@*)
}
'''
\end{lstlisting}
\chinese{...... (other examples)}
\end{tcolorbox}

    \caption{Prompt template for privacy phrase extraction with categories in step 3 for English and Chinese datasets, where <|INPUT|> is the input of the user query and <|CATEGORIES|> is the input of privacy categories.}\label{table:prompt_pipeline_step3}
\end{figure*}

\begin{figure*}[!t]
  \centering
    \begin{tcolorbox}[title=Prompts for privacy phrase deduplication (English)..]
\footnotesize
Given a user's query and privacy phrases extracted from the query, \\
Query: "<|INPUT|>"\\
Privacy Phrases: <|PHRASES|>\\
\\
Think step by step to \textcolor{flesh}{deduplicate the privacy phrases} by strictly following the given rules:\\
$\text{[}$Rule 1$\text{]}$ If privacy phrase A and privacy phrase B have the same meaning, keep phrase which is more concise and clear. \\
$\text{[}$Rule 2$\text{]}$ If privacy phrase A is part of privacy phrase B, keep phrase which has more information and is clearer. \\ 
Give your think steps. Then output the new privacy phrases after deduplication in the format of a list in json, for example:
\begin{lstlisting}[language=plainjson]
```json
[new privacy phrase 1, new privacy phrase 2]\\
'''
\end{lstlisting}
Here is an example:\\
Query: "We are happy and plan to play football."\\
Privacy Phrases: $\text{[}$"happy", "We are happy", "plan to play football", "football", "play football"$\text{]}$\\
JSON output:
\begin{lstlisting}[language=plainjson]
```json
["happy", "play football"]\\
'''
\end{lstlisting}
\end{tcolorbox}

\begin{tcolorbox}[title=Prompts for privacy phrase deduplication (Chinese).]
\footnotesize 
\chinese{给定一个用户请求（Query）和从请求中提取的隐私短语（Phrase），}\\
\chinese{用户请求：}"<|QUERY|> "\\
\chinese{隐私短语：}<|PHRASES|>\\
\\
\chinese{一步一步地思考（Think Step by Step），严格按照给定的规则\textcolor{flesh}{对隐私短语进行去重}：}\\
\chinese{$\text{[}$规则1$\text{]}$ 如果隐私短语A和隐私短语B具有相同的含义，保留更简洁明了的短语。}\\
\chinese{$\text{[}$规则2$\text{]}$ 如果隐私短语A是隐私短语B的一部分，保留信息量更大且清晰的短语。}\\
\chinese{给出你的思考步骤。然后以json列表格式输出去重后的新隐私短语，例如：}
\begin{lstlisting}[language=plainjson]
```json
[(*@\chinese{\footnotesize 新隐私短语1, 新隐私短语2}@*)]
'''
\end{lstlisting}
\chinese{以下是一个示例：}\\
\chinese{用户的请求（Query）："我们很开心，计划去踢足球。"}\\
\chinese{隐私短语：$\text{[}$'开心', '我们很开心', '计划去踢足球', '足球', '踢足球'$\text{]}$}\\
\chinese{JSON输出：}
\begin{lstlisting}[language=plainjson]
```json
[(*@\chinese{\footnotesize "开心", "踢足球"}@*)]
'''
\end{lstlisting}
\end{tcolorbox}

    \caption{Prompt template for privacy phrase deduplication in step 3 for English and Chinese dataset, where <|INPUT|> is the input of the user query and <|PHRASES|> is the input of the extracted privacy phrase.}\label{table:prompt_pipeline_step32}
\end{figure*}

\begin{figure*}[!t]
  \centering
    \begin{tcolorbox}[title=Prompts for filtering privacy phrase with rule 1 (English).]
\footnotesize
A user poses a query \\
"<|INPUT|>"\\
to you with a \textcolor{flesh}{real unique identifier}. \\
Given a phrase extracted from the query as follows:\\
Phrase: "<|PHRASE|>"\\
\\
Think step by step to \textcolor{flesh}{reason about whether the phrase satisfies the following rule}:\\
$\text{[}$Rule$\text{]}$ the phrase directly links to the user or user's related people or things in the context of the user's query. \\
Give your think steps. \textcolor{flesh}{The reasoning must depend on the direct evidence contained in the query and cannot be based on assumption}.\\
Then, output a json as follows:
\begin{lstlisting}[language=plainjson]
```json
{
    "reason": reason for your judgment,
    "judgment": true or false
}
'''
\end{lstlisting}
Here are some examples:\\
User's Query: "We plan to play football with Ross."\\
Phrase: "play football with Ross"\\
JSON Output:
\begin{lstlisting}[language=plainjson]
```json
{
    "reason": "Playing football is the user's plan that is related to the user. Ross is also related to the user.",
    "judgment": true
}
'''
\end{lstlisting}
...... (other examples)
\end{tcolorbox}

\begin{tcolorbox}[title=Prompts for filtering privacy phrase with rule 1 (Chinese).]
\footnotesize 
\chinese{用户向你提出一个请求}\\
\chinese{"<|INPUT|>"}\\
\chinese{并给定\textcolor{flesh}{用户的真实的身份标识}。}\\
\chinese{给定从请求中提取出短语如下：}\\
\chinese{短语：" <|PHRASE|> "}\\
\chinese{}\\
\chinese{一步一步地思考（Think Step by Step），\textcolor{flesh}{判断该短语是否满足以下规则}：}\\
\chinese{$\text{[}$规则$\text{]}$ 该短语在用户请求的上下文中直接与用户或与用户相关的人或事物相关联。}\\
\chinese{给出你的思考步骤。\textcolor{flesh}{推理必须依赖于请求中包含的直接证据，不能基于假设}。}\\
\chinese{然后，输出如下格式的json：}
\begin{lstlisting}[language=plainjson]
```json
{
    (*@\chinese{\footnotesize "reason" : 你判断的理由,}@*)
    (*@\chinese{\footnotesize "judgment" : true 或 false}@*)
}
'''
\end{lstlisting}
\chinese{以下是一些示例：}\\
\chinese{用户的请求："我们计划和罗斯一起踢足球。"}\\
\chinese{短语："和罗斯一起踢足球"}\\
\chinese{JSON输出：}
\begin{lstlisting}[language=plainjson]
```json
{
    (*@\chinese{\footnotesize "reason" : "踢足球是用户的计划，与用户相关。罗斯也与用户相关。",}@*)
    (*@\chinese{\footnotesize "judgment" : true}@*)
}
'''
\end{lstlisting}
\chinese{...... (other examples)}

\end{tcolorbox}

    \caption{Prompt template for filtering privacy phrase with rule 1 in step 3 for English and Chinese dataset, where <|INPUT|> is the input of the user query and <|PHRASE|> is the input of the phrase.}\label{table:prompt_pipeline_step33}
\end{figure*}

\begin{figure*}[!t]
  \centering
    \begin{tcolorbox}[title=Prompts for filtering privacy phrase with rule 2 (English).]
\footnotesize
A user poses a query \\
"<|INPUT|>"\\
to you with a \textcolor{flesh}{real unique identifier}. \\
Given a phrase extracted from the query as follows:\\
Phrase: "<|PHRASE|>"\\
\\
Think step by step to \textcolor{flesh}{reason about whether the phrase satisfies the following rule}:\\
$\text{[}$Rule$\text{]}$ the phrase should not be a general phrase that provides NO information related to the user based on the context of the user's query.\\
Give your think steps. \textcolor{flesh}{The reasoning must depend on the direct evidence contained in the query and cannot be based on assumption}.\\
Then, output a json as follows:
\begin{lstlisting}[language=plainjson]
```json
{
    "reason": reason for your judgment,
    "judgment": true or false
}
'''
\end{lstlisting}
Here are some examples:\\
User's Query: "I want to go to a place."\\
Phrases: "place"\\
JSON Output:
\begin{lstlisting}[language=plainjson]
```json
{
    "reason": "The 'place' term has no clear reference.",
    "judgment": false
}
'''
\end{lstlisting}
...... (other examples)

\end{tcolorbox}

\begin{tcolorbox}[title=Prompts for filtering privacy phrase with rule 2 (Chinese).]
\footnotesize 
\chinese{用户向你提出一个请求}\\
"<|INPUT|>"\\
\chinese{并给定\textcolor{flesh}{用户的真实的身份标识}。}\\
\chinese{给定从请求中提取出短语如下：}\\
\chinese{短语：" <|PHRASE|> "}\\
\\
\chinese{一步一步地思考（Think Step by Step），\textcolor{flesh}{判断该短语是否满足以下规则}：}\\
\chinese{$\text{[}$规则$\text{]}$ 该短语不应是一个在用户请求上下文中不提供与用户相关信息的一般短语。}\\
\chinese{给出你的思考步骤。\textcolor{flesh}{推理必须依赖于查询中包含的直接证据，不能基于假设}。}\\
\chinese{然后，输出如下格式的json：}
\begin{lstlisting}[language=plainjson]
```json
{
    "reason": (*@\chinese{\footnotesize 你判断的理由}@*),
    "judgment": true or false
}
'''
\end{lstlisting}
\chinese{以下是一些示例：}\\
\chinese{用户的请求："我想去一个地方。"}\\
\chinese{短语："地方"}\\
\chinese{JSON输出：}
\begin{lstlisting}[language=plainjson]
```json
{
    "reason": (*@\chinese{\footnotesize "'地方'一词没有明确的指代。"}@*),
    "judgment": false
}
'''
\end{lstlisting}
...... (other examples)

\end{tcolorbox}

    \caption{Prompt template for filtering privacy phrase with rule 2 in step 3 for English and Chinese dataset, where <|INPUT|> is the input of the user query and <|PHRASE|> is the input of the phrase.}\label{table:prompt_pipeline_step34}
\end{figure*}

\begin{figure*}[!t]
  \centering
    \begin{tcolorbox}[title=Prompts for privacy information annotation (English).]
\footnotesize
A user poses a query \\
"<|INPUT|>"\\
to you with \textcolor{flesh}{a real unique identifier}. \\
Given a phrase extracted from the query as follows:\\
Phrase: "<|PHRASE|>"\\
The phrase reveals \textcolor{flesh}{new privacy-sensitive information} of the user, or user related people or things to you.\\
\\
Think step by step. \\
\textcolor{flesh}{Summarize the specific privacy information} the phrase has leaked. \\
Then, output a json as follows:
\begin{lstlisting}[language=plainjson]
```json
{
    "phrase": PHRASE,
    "privacy information": PRIVACY-INFORMATION
}
'''
\end{lstlisting}
Here are some examples:\\
User's Query: "I am preparing to go back home to Minhang District, Shanghai."\\
Phrase: "Minhang District, Shanghai"\\
JSON Output:
\begin{lstlisting}[language=plainjson]
```json
{
    "phrase": "No. 800, Dongchuan Road, Minhang District, Shanghai",
    "privacy information": "The user's home address is No. 800, Dongchuan Road, Minhang District, Shanghai."
}
'''
\end{lstlisting}
...... (other examples)\\

\end{tcolorbox}

\begin{tcolorbox}[title=Prompts for privacy information annotation (Chinese).]
\footnotesize 
\chinese{用户向你提出一个请求}\\
"<|INPUT|>"\\
\chinese{并给定\textcolor{flesh}{用户的真实的身份标识}。}\\
\chinese{给定从请求中提取出短语如下：}\\
\chinese{短语：}"<|PHRASE|>"\\
\chinese{该短语向你揭示了用户或与用户相关的人或事物的，\textcolor{flesh}{新的隐私敏感信息}。}\\
\\
\chinese{一步一步地思考（Think Step by Step）。}\\
\chinese{并\textcolor{flesh}{总结短语泄露的具体隐私敏感信息}。}\\
\chinese{然后，输出如下格式的JSON：}
\begin{lstlisting}[language=plainjson]
```json
{
    "phrase": (*@\chinese{\footnotesize 短语}@*),
    "privacy information": (*@\chinese{\footnotesize 隐私信息}@*)
}
'''
\end{lstlisting}
\chinese{以下是一些示例：}\\
\chinese{用户的请求："我正在准备回家，地址是上海市闵行区东川路800号。"}\\
\chinese{短语："上海市闵行区东川路800号"}\\
\chinese{JSON输出：}
\begin{lstlisting}[language=plainjson]
```json
{
    "phrase": (*@\chinese{\footnotesize "上海市闵行区东川路800号"}@*),
    "privacy information": (*@\chinese{\footnotesize "用户的家庭地址是上海市闵行区东川路800号。"}@*)
}
'''
\end{lstlisting}
...... (other examples)

\end{tcolorbox}

    \caption{Prompt template for privacy information annotation in step 4 for English and Chinese dataset, where <|INPUT|> is the input of the user query and <|PHRASE|> is the input of the phrase.}\label{table:prompt_pipeline_step4}
\end{figure*}

\begin{figure*}[!t]
  \centering
    \begin{tcolorbox}[title=Raw data examples.]
\footnotesize
\begin{lstlisting}[language=plainjson]
"id": "ShareGPT-pKNQqpRRE1",
"conversation": [
    {
        "user": "Can you please list the operators know to work for ArcelorMittal in the world and who are not OBS?",
        "assistant": "As an AI language model, I don't have access to up-to-date information ......",
    },
    {
        "user": "David Glijer is a key target for us. How would you recommand to engage him?",
        "assistant": "When engaging with David Glijer, it's important to understand ......",
    }
]
\end{lstlisting}
\end{tcolorbox}

\begin{tcolorbox}[title=Annotated data examples of our dataset.]
\footnotesize 
\begin{lstlisting}[language=plainjson]
"id": "ShareGPT-pKNQqpRRE1",
"conversation":[
    {
        "user": "Can you please list the operators know to work for ArcelorMittal in the world and who are not OBS?",
        "assistant": "As an AI language model, I don't have access to up-to-date information ......",
        "privacy": []
    },
    {
        "user": "David Glijer is a key target for us. How would you recommand to engage him?",
        "assistant": "When engaging with David Glijer, it's important to understand ......",
        "privacy": [
            {
                "phrase": "David Glijer",
                "privacy information": "David Glijer is identified as a key target of interest, suggesting a strategic focus in a business context."
            },
            {
                "phrase": "key target for us",
                "privacy information": "David Glijer is an important person of interest or focus for the user's objectives, implying a strategic or professional engagement."
            },
            {
                "phrase": "engage him",
                "privacy information": "The user has plans to strategically engage or interact with David Glijer, indicating he is a subject of interest."
            }
        ]
    }
]
\end{lstlisting}
\end{tcolorbox}

    \caption{Data samples with privacy phrase and privacy information. The samples come from one raw dialogue.}\label{table:dataset_example}
\end{figure*}

\begin{figure*}[!t]
  \centering
    \begin{tcolorbox}[title=Prompts for zero-shot generation (English).]
\footnotesize
\textcolor{flesh}{Determine whether the query leaks the privacy information} of the user, or user related people or things.\\
The privacy information includes any opinions, concerns, preferences, activities, intentions, or any other information that could\\
be considered private or sensitive based on the context of the query.\\
Finally, output your answer in JSON. \\
Here is the format:
\begin{lstlisting}[language=plainjson]
```json
{
    "judgment": true or false
}
'''
\end{lstlisting}
The user's QUERY is \\
"<|QUERY|>"
\end{tcolorbox}

\begin{tcolorbox}[title=Prompts for zero-shot generation (Chinese).]
\footnotesize 
\chinese{\textcolor{flesh}{判断用户请求是否泄露}了用户、用户相关人员或事物的隐私敏感信息。}\\
\chinese{隐私敏感信息包括用户的观点、偏好、活动、意图或根据请求上下文可能被认为是私人或敏感的信息。}\\
\chinese{请一步一步思考（Think step-by-step）。}\\
\chinese{给出理由并在最后，以JSON格式输出答案。}\\
\chinese{以下是格式：}  
\begin{lstlisting}[language=plainjson]
```json
{
    "judgment": true or false
}
'''
\end{lstlisting}
\chinese{用户的请求是：}\\
"<|QUERY|>"
\end{tcolorbox}

    \caption{Prompt template for zero-shot generation of privacy leakage classification task in English and Chinese dataset, where <|INPUT|> is the input of the user query.}\label{table:prompt_zero_query}
\end{figure*}

\begin{figure*}[!t]
  \centering
    \begin{tcolorbox}[title=Prompts for zero-shot generation (English).]
\footnotesize
\textcolor{flesh}{Extract informative phrases} from the query that reveal the privacy information of the user, or user related people or things.\\
The privacy information includes any opinions, concerns, preferences, activities, intentions, or any other information that could be considered private or sensitive based on the context of the query.\\
\textcolor{flesh}{The phrase should be relevant to the user or the user’s related people or things based on the context of the query.}\\
\textcolor{flesh}{The phrase should not be a general term.}\\
You should think step-by-step.\\
Finally, output your answer in JSON. If there is no phrase extracted, the output JSON list should be empty.\\
Here is the format:
\begin{lstlisting}[language=plainjson]
```json
["phrase1", "phrase2", ...]
'''
\end{lstlisting}
The user's QUERY is \\
"<|QUERY|>"
\end{tcolorbox}

\begin{tcolorbox}[title=Prompts for zero-shot generation (Chinese).]
\footnotesize 
\chinese{提取用户请求中，揭示了用户、用户相关人员或事物的隐私敏感信息的短语。}\\
\chinese{隐私敏感信息包括用户的观点、偏好、活动、意图或根据请求上下文可能被认为是私人或敏感的信息。}\\
\chinese{\textcolor{flesh}{提取的短语应当在请求的上下文中体现出与用户或用户相关的人员或事物的相关性}。}\\
\chinese{\textcolor{flesh}{隐私短语应当有具体的指代，而不是一个泛指的词汇。}}\\
\chinese{请一步一步思考（Think step-by-step）。}\\
\chinese{给出理由并在最后，以JSON格式输出答案。如果没有提取到任何短语，则输出的JSON列表应为空。}\\
\chinese{以下是格式：}  
\begin{lstlisting}[language=plainjson]
```json
[(*@\chinese{\footnotesize "短语1", "短语2", ...}@*)]
'''
\end{lstlisting}
\chinese{用户的请求是：}\\
"<|QUERY|>"
\end{tcolorbox}

    \caption{Prompt template for zero-shot generation of privacy phrase extraction task in English and Chinese dataset, where <|INPUT|> is the input of the user query.}\label{table:prompt_zero_phrase}
\end{figure*}

\begin{figure*}[!t]
  \centering
    \begin{tcolorbox}[title=Prompts for zero-shot generation (English).]
\footnotesize
Extract informative phrases from the query that \textcolor{flesh}{reveal the privacy information of the user, or user related people or things}.\\
The privacy information includes any opinions, concerns, preferences, activities, intentions, or any other information that could\\
be considered private or sensitive based on the context of the query.\\
The phrase should be relevant to the user or the user’s related people or things based on the context of the query.\\
Then for each extracted phrase, \textcolor{flesh}{summarize the specific privacy information} the phrase has leaked.\\
You should think step-by-step.\\
Finally, output your answer in JSON. If there is no phrase extracted, the output JSON list should be empty.\\
Here is the format:
\begin{lstlisting}[language=plainjson]
```json
[
    {
        "phrase": PHRASE-1,
        "privacy information": PRIVACY-INFORMATION-1
    }
]
'''
\end{lstlisting}
The user's QUERY is \\
"<|QUERY|>"
\end{tcolorbox}
\begin{tcolorbox}[title=Prompts for zero-shot generation (Chinese).]
\footnotesize 
\chinese{提取用户请求中，\textcolor{flesh}{揭示了用户、用户相关人员或事物的隐私敏感信息}的短语。}\\
\chinese{隐私敏感信息包括用户的观点、偏好、活动、意图或根据请求上下文可能被认为是私人或敏感的信息。}\\
\chinese{提取的短语应当在请求的上下文中体现出与用户或用户相关的人员或事物的相关性。}\\
\chinese{然后，对于每个提取的短语，\textcolor{flesh}{总结该短语泄露的具体隐私信息}。}\\
\chinese{请一步一步思考（Think step-by-step）。}\\
\chinese{给出理由并在最后，以JSON格式输出答案。如果没有提取到任何短语，则输出的JSON列表应为空。}\\
\chinese{以下是格式：}  
\begin{lstlisting}[language=plainjson]
```json
[
    {
        "phrase": (*@\chinese{\footnotesize "短语",}@*)
        "privacy information": (*@\chinese{\footnotesize "具体的隐私信息"}@*)
    }
]
'''
\end{lstlisting}
\chinese{用户的请求是：}\\
"<|QUERY|>"
\end{tcolorbox}

    \caption{Prompt template for zero-shot generation of privacy information summarization task in English and Chinese dataset, where <|INPUT|> is the input of the user query.}\label{table:prompt_zero_info}
\end{figure*}

\begin{figure*}[!t]
  \centering
    \begin{tcolorbox}[title=Prompts for in-context learning (English).]
\footnotesize
\textcolor{flesh}{Determine whether the query leaks the privacy information} of the user, or user related people or things.\\
The privacy information includes any opinions, concerns, preferences, activities, intentions, or any other information that could\\
be considered private or sensitive based on the context of the query.\\
Finally, output your answer in JSON. \\
\\
Here are some examples:\\
<|CASE|>End of examples.\\
\\
The user's current Query is \\
"<|QUERY|>"
\end{tcolorbox}

\begin{tcolorbox}[title=Prompts for in-context learning (Chinese).]
\footnotesize 
\chinese{\textcolor{flesh}{判断用户请求是否泄露}了用户、用户相关人员或事物的隐私敏感信息。}\\
\chinese{隐私敏感信息包括用户的观点、偏好、活动、意图或根据请求上下文可能被认为是私人或敏感的信息。}\\
\chinese{请一步一步思考（Think step-by-step）。}\\
\chinese{给出理由并在最后，以JSON格式输出答案。}\\
\\
\chinese{以下是一些例子：}\\
\chinese{<|CASE|>例子结束。}\\
\\
\chinese{用户当前的请求是：}\\
"<|QUERY|>"
\end{tcolorbox}

    \caption{Prompt template for in-context learning of privacy leakage classification task in English and Chinese dataset, where <|INPUT|> is the input of the user query and <|CASE|>is structured examples from the training set.}\label{table:prompt_icl_query}
\end{figure*}

\begin{figure*}[!t]
  \centering
    \begin{tcolorbox}[title=Prompts for in-context learning (English).]
\footnotesize
\textcolor{flesh}{Extract informative phrases} from the query that reveal the privacy information of the user, or user related people or things.\\
The privacy information includes any opinions, concerns, preferences, activities, intentions, or any other information that could be considered private or sensitive based on the context of the query.\\
\textcolor{flesh}{The phrase should be relevant to the user or the user’s related people or things based on the context of the query.} \\
\textcolor{flesh}{The phrase should not be a general term.}\\
You should think step-by-step.\\
Finally, output your answer in JSON. If there is no phrase extracted, the output JSON list should be empty.\\
\\
Here are some examples:\\
<|CASE|>End of examples.\\
\\
The user's current Query is \\
"<|QUERY|>"
\end{tcolorbox}

\begin{tcolorbox}[title=Prompts for in-context learning (Chinese).]
\footnotesize 
\chinese{提取用户请求中，揭示了用户、用户相关人员或事物的隐私敏感信息的短语。}\\
\chinese{隐私敏感信息包括用户的观点、偏好、活动、意图或根据请求上下文可能被认为是私人或敏感的信息。}\\
\chinese{\textcolor{flesh}{提取的短语应当在请求的上下文中体现出与用户或用户相关的人员或事物的相关性。}}\\
\chinese{\textcolor{flesh}{隐私短语应当有具体的指代，而不是一个泛指的词汇。}}\\
\chinese{请一步一步思考（Think step-by-step）。}\\
\chinese{给出理由并在最后，以JSON格式输出答案。如果没有提取到任何短语，则输出的JSON列表应为空。}\\
\\
\chinese{以下是一些例子：}\\
\chinese{<|CASE|>例子结束。}\\
\\
\chinese{用户当前的请求是：}\\
"<|QUERY|>"
\end{tcolorbox}

    \caption{Prompt template for in-context learning of privacy phrase extraction task in English and Chinese dataset, where <|INPUT|> is the input of the user query and <|CASE|>is structured examples from the training set.}\label{table:prompt_icl_phrase}
\end{figure*}

\begin{figure*}[!t]
  \centering
    \begin{tcolorbox}[title=Prompts for in-context learning (English).]
\footnotesize
Extract informative phrases from the query that \textcolor{flesh}{reveal the privacy information of the user, or user related people or things}.\\
The privacy information includes any opinions, concerns, preferences, activities, intentions, or any other information that could\\
be considered private or sensitive based on the context of the query.\\
The phrase should be relevant to the user or the user’s related people or things based on the context of the query.\\
Then for each extracted phrase, \textcolor{flesh}{summarize the specific privacy information} the phrase has leaked.\\
You should think step-by-step.\\
Finally, output your answer in JSON. If there is no phrase extracted, the output JSON list should be empty.\\
\\
Here are some examples:\\
<|CASE|>End of examples.\\
\\
The user's current Query is \\
"<|QUERY|>"
\end{tcolorbox}

\begin{tcolorbox}[title=Prompts for in-context learning (Chinese).]
\footnotesize 
\chinese{提取用户请求中，\textcolor{flesh}{揭示了用户、用户相关人员或事物的隐私敏感信息}的短语。}\\
\chinese{隐私敏感信息包括用户的观点、偏好、活动、意图或根据请求上下文可能被认为是私人或敏感的信息。}\\
\chinese{提取的短语应当在请求的上下文中体现出与用户或用户相关的人员或事物的相关性。}\\
\chinese{然后，对于每个提取的短语，\textcolor{flesh}{总结该短语泄露的具体隐私信息}。}\\
\chinese{请一步一步思考（Think step-by-step）。}\\
\chinese{给出理由并在最后，以JSON格式输出答案。如果没有提取到任何短语，则输出的JSON列表应为空。}\\
\\
\chinese{以下是一些例子：}\\
\chinese{<|CASE|>例子结束。}\\
\\
\chinese{用户当前的请求是：}\\
"<|QUERY|>"
\end{tcolorbox}

    \caption{Prompt template for in-context learning of privacy information summarization task in English and Chinese dataset, where <|INPUT|> is the input of the user query and <|CASE|>is structured examples from the training set.}\label{table:prompt_icl_info}
\end{figure*}

\begin{figure*}[!t]
  \centering
    \begin{tcolorbox}[title=Instruction for English datasets.]
\footnotesize
\textbf{\textit{System}}: \\ 
Judge \textcolor{flesh}{whether the query that reveals any privacy-sensitive information of the user, or user related people or things}. \\
\textbf{\textit{Message}}: \\
Query: "<|INPUT|>"
\end{tcolorbox}

\begin{tcolorbox}[title=Instruction for Chinese datasets.]
\footnotesize 
\textbf{\textit{System}}: \\ 
\chinese{判断Query\textcolor{flesh}{是否泄漏了用户、用户相关人员或事物的隐私敏感信息}。}\\
\textbf{\textit{Message}}: \\ 
\chinese{Query: "<|INPUT|>"}
\end{tcolorbox}

    \caption{Instructions for training models for privacy leakage classification task, where <|INPUT|> is the input of the user query.}\label{table:prompt_train_query}
\end{figure*}

\begin{figure*}[!t]
  \centering
    \begin{tcolorbox}[title=Instruction for English datasets.]
\footnotesize
\textbf{\textit{System}}: \\ 
\textcolor{flesh}{Extract informative phrases} from the query that \textcolor{flesh}{reveal the privacy-sensitive information of the user, or user related people or things.} \\
\textbf{\textit{Message}}: \\
Query: "<|INPUT|>"
\end{tcolorbox}

\begin{tcolorbox}[title=Instruction for Chinese datasets.]
\footnotesize 
\textbf{\textit{System}}: \\ 
\chinese{从Query中提取出\textcolor{flesh}{能够揭示用户、用户相关人员或事物的隐私敏感信息的短语}。}\\
\textbf{\textit{Message}}: \\ 
\chinese{Query: "<|INPUT|>"}
\end{tcolorbox}

    \caption{Instructions for training models for privacy phrase extraction task, where <|INPUT|> is the input of user query.}\label{table:prompt_train_phrase}
\end{figure*}

\begin{figure*}[!t]
  \centering
    \begin{tcolorbox}[title=Instruction for English datasets.]
\footnotesize
\textbf{\textit{System}}: \\ 
Extract informative phrases from the query that reveal the privacy-sensitive information of the user, or user related people or things. \\
Then for each extracted phrase, \textcolor{flesh}{summarize the specific privacy information} the phrase has leaked. \\
\textbf{\textit{Message}}: \\
Query: "<|INPUT|>"
\end{tcolorbox}

\begin{tcolorbox}[title=Instruction for Chinese datasets.]
\footnotesize 
\textbf{\textit{System}}: \\ 
\chinese{从Query中提取出能够揭示用户、用户相关人员或事物的隐私敏感信息的短语。}\\
\chinese{然后，针对每个提取的短语，总结该短语\textcolor{flesh}{泄露的具体隐私信息}。}\\
\textbf{\textit{Message}}: \\ 
Query: "<|INPUT|>"
\end{tcolorbox}

    \caption{Instructions for training models for privacy information summarization task, where <|INPUT|> is the input of the user query.}\label{table:prompt_train_info}
\end{figure*}

\end{document}